\RequirePackage{fix-cm}
\documentclass[smallcondensed]{svjour3} 

\smartqed  
\usepackage{appendix}
\usepackage{amsmath}
\usepackage{graphicx}
\usepackage{lineno}
\usepackage{array}
\usepackage{longtable}
\usepackage[numbers]{natbib}
\usepackage{booktabs}
\usepackage{amssymb}
\usepackage{subfig}
\usepackage{xcolor}
\usepackage{tabularx}
\usepackage{booktabs}
\usepackage{algorithm}
\usepackage{amsmath,amssymb}
\usepackage{ragged2e}
\usepackage{tabularx}
\usepackage{color}
\usepackage{algorithmic}

\usepackage{collcell}
\usepackage{hhline}
\usepackage{pgf}
\usepackage{multirow}
\usepackage{enumitem}
\usepackage[misc,geometry]{ifsym} 
\usepackage{lipsum}
\usepackage{lscape}
\usepackage{pifont}
\newcommand{\cmark}{\ding{51}}%
\newcommand{\xmark}{\ding{55}}
\newcommand*\patchAmsMathEnvironmentForLineno[1]{%
\expandafter\let\csname old#1\expandafter\endcsname\csname #1\endcsname
\expandafter\let\csname oldend#1\expandafter\endcsname\csname end#1\endcsname
\renewenvironment{#1}%
{\linenomath\csname old#1\endcsname}%
{\csname oldend#1\endcsname\endlinenomath}}%
\newcommand*\patchBothAmsMathEnvironmentsForLineno[1]{%
\patchAmsMathEnvironmentForLineno{#1}%
\patchAmsMathEnvironmentForLineno{#1*}}%
\AtBeginDocument{%
\patchBothAmsMathEnvironmentsForLineno{equation}%
\patchBothAmsMathEnvironmentsForLineno{align}%
\patchBothAmsMathEnvironmentsForLineno{flalign}%
\patchBothAmsMathEnvironmentsForLineno{alignat}%
\patchBothAmsMathEnvironmentsForLineno{gather}%
\patchBothAmsMathEnvironmentsForLineno{multline}%
}

\begin{document}
\title{
Recent Advances in Scene Image Representation and Classification
}


\author{Chiranjibi Sitaula* \and Tej Bahadur Shahi \and Faezeh Marzbanrad \and Jagannath Aryal
}


\institute{
             Corresponding Author (*C. Sitaula) \and F. Marzbanrad \at
              Department of Electrical and Computer Systems Engineering\\ Monash University\\
              Wellington Rd, Clayton VIC 3800, Australia\\
              \email{chiranjibi.sitaula@monash.edu}  \\
              \and 
              TB Shahi \at
               School of Engineering and Technology\\
               Central Queensland University, Rockhampton, QLD, 4701, Australia\\
               and \\
              Central Department of Computer Science and Information Technology (CDCSIT)\\
              Tribhuvan University\\
               TU Rd, Kirtipur 44618, Kathmandu, Nepal \\
                \and
              J. Aryal \at
              Department of Infrastructure Engineering \\
              The University of Melbourne\\
              Parkville VIC 3010, Australia\\
}

\date{Received: DD Month YEAR / Accepted: DD Month YEAR}

\maketitle

\begin{abstract}
With the rise of deep learning algorithms nowadays, scene image representation methods have achieved a significant performance boost in classification. However, the performance is still limited because the scene images are mostly complex having higher intra-class dissimilarity and inter-class similarity problems. To deal with such problems, there have been several methods proposed in the literature with their advantages and limitations. A detailed study of previous works is necessary to understand their advantages and disadvantages in image representation and classification problems. 
In this paper, we review the existing scene image representation methods that are being widely used for image classification. For this, we, first, devise the taxonomy using the seminal existing methods proposed in the literature to this date {using deep learning (DL)-based, computer vision (CV)-based and search engine (SE)-based methods}. Next, we compare their performance both qualitatively (e.g., quality of outputs, pros/cons, etc.) and quantitatively (e.g., accuracy). Last, we speculate on the prominent research directions in scene image representation tasks using {keyword growth and timeline analysis.} 
Overall, this survey provides in-depth insights and applications of recent scene image representation methods under three different methods.

\end{abstract}

\section{Introduction}

Scene image analytics (e.g., scene representation, classification, clustering, etc.) is a highly-researched topic owing to its strong connection to recent technologies such as sensors, video cameras, robotics, and the internet of things (IoT) \cite{sitaula2019indoor}.
It also has an association with other sectors such as hyperspectral image analytics \cite{shadman2019uncertainty}, satellite image analytics \cite{neupane2021deep}, climate image analytics \cite{dutta2013deep}, and so on.
The image representation methods for each of them are dependent on the nature of the images; therefore, we need to adopt the appropriate feature extraction methods for their representation accordingly \cite{sitaula2019unsupervised}. To perform such tasks, researchers have extended their works from very basic levels that use traditional computer vision-based methods to more sophisticated levels that use recent deep learning-based methods in addition to search engine-based methods.

Initially, researchers mostly preferred to use the traditional Computer Vision (CV)-based methods until 2014 for the scene image representation tasks. This is because Deep Learning (DL) models {did not} flourish at that time and traditional CV-based methods dominated scene representation tasks.
Later on, DL-based methods, which {originated} in 1943 \cite{mcculloch1943logical}, have been dominant in the computer vision community from 2014 until now, particularly for scene image representation and classification \cite{sitaula2019indoor}.
Recently, to tackle the weaknesses of visual information achieved from either traditional CV-based methods or DL-based methods, in 2019, researchers proposed new methods based on the Search Engine (SE) to capture the contextual information for the scene image representation tasks, which are also called SE-based methods \cite{sitaula2019tag}.

\begin{table*}[b]

\vspace{0mm}
\centering
\begin{tabular}{p{2.5cm} p{1.5cm} p{1.5cm} p{1.5cm} p{1.5cm} p{1cm}}
\toprule
Questions &Wei et al. \cite{wei2016visual}& Anu et al. \cite{anu2016survey} & Singh et al. \cite{singh2017image}& Xie et al. \cite{xie2020scene}& {\bf Ours}\\
\midrule
Traditional CV-based methods? &\cmark&\cmark&\cmark&\cmark&\cmark\\
\hline
Latest DL-based methods? &\xmark&\xmark&\xmark&\cmark&\cmark \\
\hline
SE-based methods? &\xmark&\xmark& \xmark& \xmark&\cmark \\
 \hline
{Trend and keyword growth analysis?} &\xmark&\xmark&\xmark&\xmark&\cmark \\
\hline
\end{tabular}
\caption{Comparison of our work with existing works}
\label{tab:comparison_of_survey_papers}
\end{table*} 

Because of such predominant growth and application of such methods, it has been challenging to explore the potential of each of them. Therefore, a survey study is crucial, not only to explore the surging potentials but also to help understand the application areas,{ research trends, and developments.}
{Some recent review works related to scene image representation are summarised below, whereas the summary is {reported} in Table \ref{tab:comparison_of_survey_papers}.}

\begin{itemize}
    \item[(i)] Wei et al. \cite{wei2016visual} studied the traditional feature extraction methods using empirical analysis, when the DL-based methods were not dominant, which helped understand the efficacy of traditional feature extraction methods for scene image representation. {In addition, they perform an empirical study of such methods on four benchmark datasets.}
    {However, they explain a limited DL-methods for scene image representation, which lacks in-depth elaboration of recent DL-methods in this domain. }
    
    \item[(ii)] Anu et al. \cite{anu2016survey} discussed the traditional CV-based methods to extract the image features, which shed light on the applicability of different CV-based methods for scene image representation during that time.
    However, their study does not classify the traditional CV-based methods in-detail. 
    
    \item[(iii)] Singh et al. \cite{singh2017image} presented a review of recent methods of scene representation, including DL-based methods, which provided a great promise of DL-based methods for scene image representation. They categorised the range of methods into three broad categories. However, their study limits recent advances of DL-based methods in this domain.
    
    \item[(iv)] Xie et al. \cite{xie2020scene} discussed the recent DL-based methods and traditional CV-based methods for scene representation, which not only carried out an in-depth study of each of them but also underscored the efficacy of DL-based methods against other methods for the scene image representation. However, their study has two main limitations. First, semantic approaches (e.g., SE-based methods) that have been gaining popularity recently are not included in their study. Second, their study lacks the comparative study of traditional CV-based methods, DL-based methods, and SE-based methods.
    
\end{itemize}


To bridge the gaps in existing survey works, we study the recent and existing methods used in scene recognition and analyse them under their appropriate taxonomy using both qualitative and quantitative analysis. {In addition, we present the ongoing research trends in scene image representation.}

The main {\bf contributions} in this paper are as follows:
\begin{enumerate}
    
    \item[(i)] We perform a detailed review of the existing and recent scene image representation methods for classification using a comprehensive taxonomy.
    
    \item[(ii)] We analyse the existing scene representation methods qualitatively and quantitatively. For quantitative analysis, we use a statistical approach, particularly box-plot analysis, across the performance measures, whereas, for qualitative analysis, we take the help of the pros/cons of methods.
    
    \item[(iii)] Based on the pros and cons of the existing methods, we point out the potential directions of scene image representation and classification.
    \item[(iv)] {We reveal the trend and keyword growth analysis in scene image representation area.}

\end{enumerate}

The rest of the paper is organised as follows. Sec. \ref{survey_method} explains the process used to retrieve the papers for review. Similarly, Sec. \ref{background} provides the basic concepts used in the scene representation, and Sec. \ref{taxonomy} categorises the existing methods into three broad categories with their explanation. Sec. \ref{datasets_performance} explains the datasets used in the scene representation and details the comparative study of the existing methods and Sec. \ref{discussion} discusses the overall methods and suggests the possible directions. Finally, Sec. \ref{conclusion} concludes the paper with final remarks.

\section{Survey Method}
\label{survey_method}

\begin{figure}[tb]
    \centering
    \includegraphics[width=\textwidth, height=80mm,keepaspectratio]{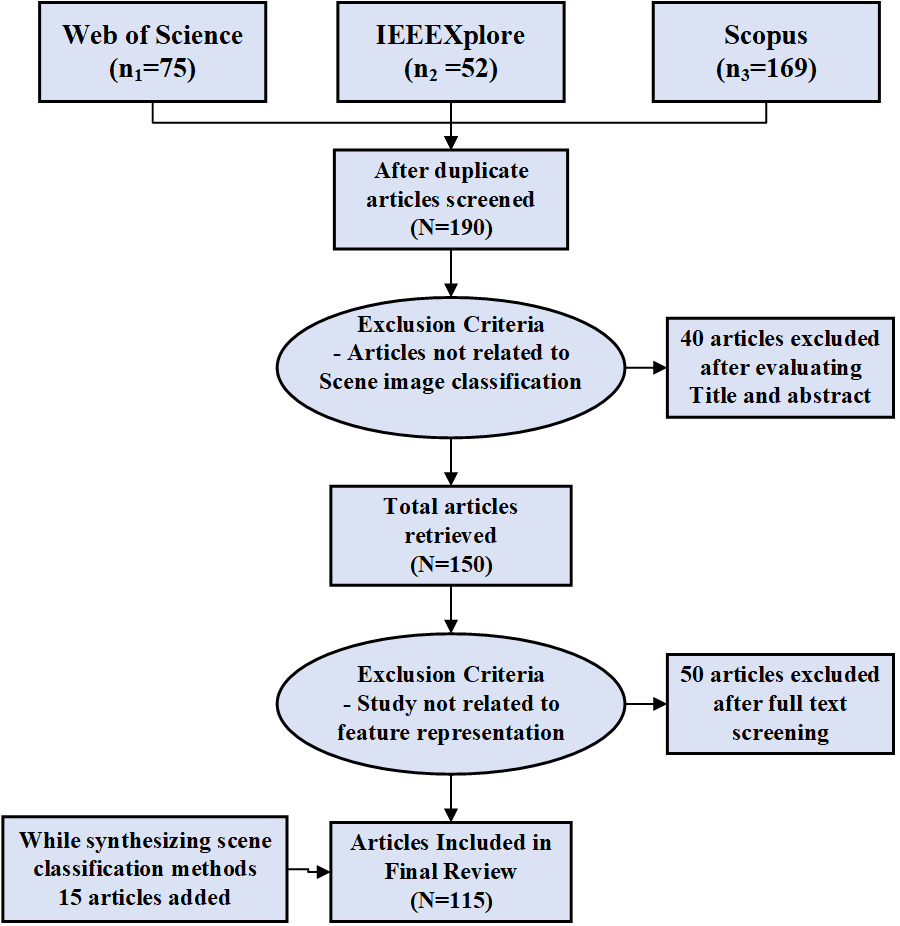} 
    \caption{Step-wise procedure to retrieve the articles reviewed in this survey.}
    \label{fig:pipeline_survey_method}
\end{figure}

In this section, we outline the procedure to retrieve the papers for review. We follow a systematic procedure to collect the papers for review. For this, we first search three popular databases: IEEE Xplore, Scopus, and Web of Science with the search string: "Scene Image OR {Place}" AND "Representation" AND "Classification". With this, we find 52, 169, and 75 articles with IEEE Xplore, Scopus, and Web of Science, respectively (Accessed date: 2022/11/10). After screening the title, abstract, author keywords, and full text, we end up collecting 100 articles. {In addition to the searching method}, we also collect {15} related articles using a snowballing technique. Last, a total of {115} articles are included for final review, including both scene representation methods and their related articles. The detailed pipeline of our survey method is presented in Fig. \ref{fig:pipeline_survey_method}. 

\section{Background}
\label{background}

Here, we explain the fundamental concepts used by existing methods in scene representation problems mostly. 

\subsection{Scale Invariant Feature Transform (SIFT)}

SIFT feature extraction algorithm, which was published in Lowe et al. \cite{lowe1999object}, extracts the features based on the local sense of the image. This algorithm is mainly used for object recognition, gesture recognition, video tracking, etc.;  however, it has also been used in scene representation problems \cite{lazebnik2006beyond}. 
It is a complex algorithm, which follows four steps to extract the descriptor: a) Scale-space detection, b) Key points localization, c) Orientation assignment, and d) Key points descriptor. 

At first, to detect the key points in scale-space detection, multiple-scaled images are created and scale filtering is performed. For this, Laplacian of Gradient (LoG) could be used as a blob detection in each scale. However, since the LoG is a little bit costly, the Difference of Gaussian (DoG) is used in SIFT descriptor. The DoG is obtained by the difference of Gaussian blurring of an image with two differences $\sigma$, such as $\sigma$ and $k\sigma$. 
Once the DoGs are achieved using such an approach, local maxima are found by searching the image with different scales and spaces. Local maxima are the potential key points of the corresponding image.

After the identification of potential key points in scale-space detection, the second step is to refine them for accurate results. For this, the Taylor series expansion algorithm \cite{moller1997evaluation} is used to get a more accurate location of local maxima in addition to the contrast threshold approach. With the help of the contrast threshold, we choose those extrema that have less than the threshold (e.g., 0.03), which can be chosen empirically. Furthermore, DoG exploits the edge information, which needs to be removed. Thus, the Harris corner detector is used to detect them and another threshold, called the edge threshold, is used to filter them out.
With the help of such an approach, the extrema with low-intensity and edge key points are removed, thereby preserving only strong intensity key points.

Next, the third step provides the in-variance to the extracted key points. In this step, orientation is assigned to each key point, where the neighborhood is considered into account around each key point depending on the scale, gradient, and direction. In this way, an orientation histogram is created with 36 bins covering 360 degrees. The highest peak of the histogram is taken and a peak below 80\% is discarded. 

Finally, the descriptor is created by taking the window of 16 $\times$ 16 neighborhood around the key points. Such a neighborhood is divided into 16 sub-blocks of 4 $\times$ 4, where for each sub-block, an orientation histogram of having 8 bins is constructed. This results in 128 bins in total for each key point. In this way, SIFT descriptor is created.

\subsection{Histogram of Gradient (HoG)}
HoG features also focus on the local sense, that is the gradient in the images. This concept was brought by Dalal et al. \cite{dalal2005histograms}. It was initially used to detect the objects in the image; however, it has been used in scene recognition problems these days \cite{xie2018improved}. To extract the HoG descriptors, we follow three steps: computation of gradient, orientation binning, and descriptor blocks.

First, the gradient values are calculated for an image. Specifically, this step utilizes filtering the color or intensity data of the image using two kernels such as [-1,0,1] and [-1,0,1]$^T$.
Next, the histograms of cells are constructed. The structure of the cells can be either rectangular or radial and the histogram channels are spread over 0 to 180 or 0 to 360 degrees depending on the unsigned or signed gradient, respectively. Then, these histograms are normalized.
Last, the HoG descriptor is obtained by the concatenation of all normalized cell histograms. Such blocks generally overlap, which means that each contributes more than once to form the descriptor.

\subsection{CENsus TRansform hISTogram (CENTRIST) and Multi-channel CENTRIST (mCENTRIST) }
The CENTRIST descriptor captures the structural detail of the image with the help of local structural detail. For this, spatial geometric information is utilized. To achieve such spatial information, it uses CT (Census Transform) values as its basic component. CT value is defined as the non-parametric local transform established to show the association between the intensity values \cite{zabih1994non}. To show the association in CT values, the intensity values are set to 0 if it is greater than the center value and set to 1 otherwise (Eq. \eqref{eq:ct}).
Here, CT values (e.g, CT=224 for 20 in Eq. \eqref{eq:ct}) are calculated based on its 8 neighbouring intensity values. Finally, all the CT values are collected and constructed in the histogram to form the CENTRIST descriptor.

\begin{equation}
\begin{pmatrix}
 10 &  20& 30\\ 
 10&  \bf 20& 30\\ 
 10&  20& 30
\end{pmatrix}
\overset{}{\Rightarrow} 
\begin{pmatrix}
 1 &  1& 0\\ 
 1&  & 0\\ 
 1&  1& 0
\end{pmatrix} 
\overset{}{\Rightarrow} (11010110)_2\\
\overset{}{\Rightarrow}224
\label{eq:ct}
\end{equation}

Furthermore, the mCENTRIST \cite{xiao_mcentrist:_2014} descriptor is the multi-channel CENTRIST descriptor, which is developed to overcome the weaknesses of CENTRIST. CENTRIST has mainly two weaknesses: first, it extracts the descriptor using a single channel; second, its descriptor size is larger. To overcome the weaknesses of CENTRIST, mCENTRIST uses complementary information using two or multiple channels, which improves the performance. Similarly, with the help of the Census Transform pyramid, they can reduce the size of the descriptor significantly.

\subsection{Oriented Texture Curves}
To achieve the OTC \cite{margolin2014otc} descriptor, we need to perform three main steps. First, we need to sample the patches along the dense grid of the image. Next, each patch is represented by the curve, where each curve is based on a certain curve descriptor, that is texture-based and rotation sensitive. Note that for the texture-based descriptor, we use the HoG descriptor in the method. Last, such descriptor is concatenated and normalized to achieve the OTC descriptor.

\subsection{Deep features}
Deep features, which are the deep visual representation of the image, are extracted using various intermediate layers of {deep learning model such as VGG16 \cite{sitaula2020scene}}. Deep features achieved from different layers provide different kinds of information (e.g., foreground, background, etc.), which can be used to describe the various contents present in the image \cite{sitaula2020scene, sitaula2019unsupervised, sitaula2020content,sitaula2020hdf}.
Moreover, deep features represent the image at a higher order; therefore, it can discriminate such images more accurately than traditional computer vision-based descriptors such as SIFT, HoG, and so on.

\subsection{Word embedding}
Descriptors can also be achieved using the word embedding form from the pre-trained models \cite{mikolov2013efficient,pennington2014glove,bojanowski2017enriching}. Such descriptors, which are popular in Natural Language Processing (NLP) \cite{shahi2021natural}, have been used to extract the contextual information using tags/tokens representing the scene image \cite{sitaula2019tag}.
There are basically three types of word embedding used in NLP tasks, which have also been used in image processing to capture contextual information. They are Word2Vec \cite{mikolov2013efficient}, GloVe \cite{pennington2014glove}, and fastText \cite{bojanowski2017enriching}.


\subsection{Sparse coding}
Sparse coding yields the sparse representation of the input image based on the dictionary learning method. Based on the training images, a dictionary is constructed at first. Then, with the help of such a dictionary and its optimization, sparse representation to attain the final encoded features representing the image. This algorithm is popular in scene representation \cite{nascimento_robust_2017}.

\subsection{Bag of visual words}
The bag of Visual Words (BoVW) encoding method is a slight variation of the bag of words (BoW) approach, which is quite popular in the Natural Language Processing (NLP) domain mostly. The BoVW method is invariant to scale and orientation, which is helpful to achieve better performance irrespective of the different resolutions and orientations of scene images.  
This method has been used widely in the computer vision domain nowadays \cite{lazebnik2006beyond}. To employ the BoVW in computer vision, the frequencies of visual words are considered, unlike the BoW approach.

\subsection{Fisher vectors}
To avoid the problem of sparsity and higher dimensionality problem in BoVW, the concept of Fisher vectors (FV) \cite{sanchez2013image}, which adopt the Fisher Kernel (the compact and dense representation), has been used. Specifically, the Fisher Vector (FV) is the general Fisher kernel, which is obtained by pooling local image features. For this, it stores the mean and covariance deviation vectors per component k of the Gaussian Mixture Model (GMM) in addition to each element of the local descriptor.

\subsection{Locally-constrained Linear coding (LLC)}
In LLC, each descriptor is projected to locality constraints using a local co-ordinate system and then, the projected co-ordinates are integrated using max-pooling operation, which results in the final representation \cite{li2012reference}. This encoding is also popular to attain the fixed-sized features for the scene image representation. 

\subsection{Principal Component Analysis}
Principal Component Analysis (PCA) \cite{ringner2008principal} has been used to reduce the dimension of the higher feature size. However, since it can provide fixed-sized features, it has also been used as an encoding algorithm. 
PCA extracts the orthogonal set of variables, which are called principal components (PCs). Based on those PCs, we achieve the reduced and fixed size of features. In the literature on scene image representation problems, this method has been used to reduce the deep feature size before the classification takes place \cite{sitaula2020scene}.

\subsection{Threshold-based histogram}
This is an approach, where the fixed-sized features are constructed using the threshold operation to increment each bin of the histogram. Although this approach is computationally expensive, it can capture discriminating information. In scene representation, this approach has been used in SE-based algorithms to attain the feature vector representing the textual information \cite{sitaula2019tag}.

\section{Taxonomy of scene image representation methods}
\label{taxonomy}

\begin{figure}
    \centering
    \includegraphics[width=0.98\textwidth, height=150mm, keepaspectratio]{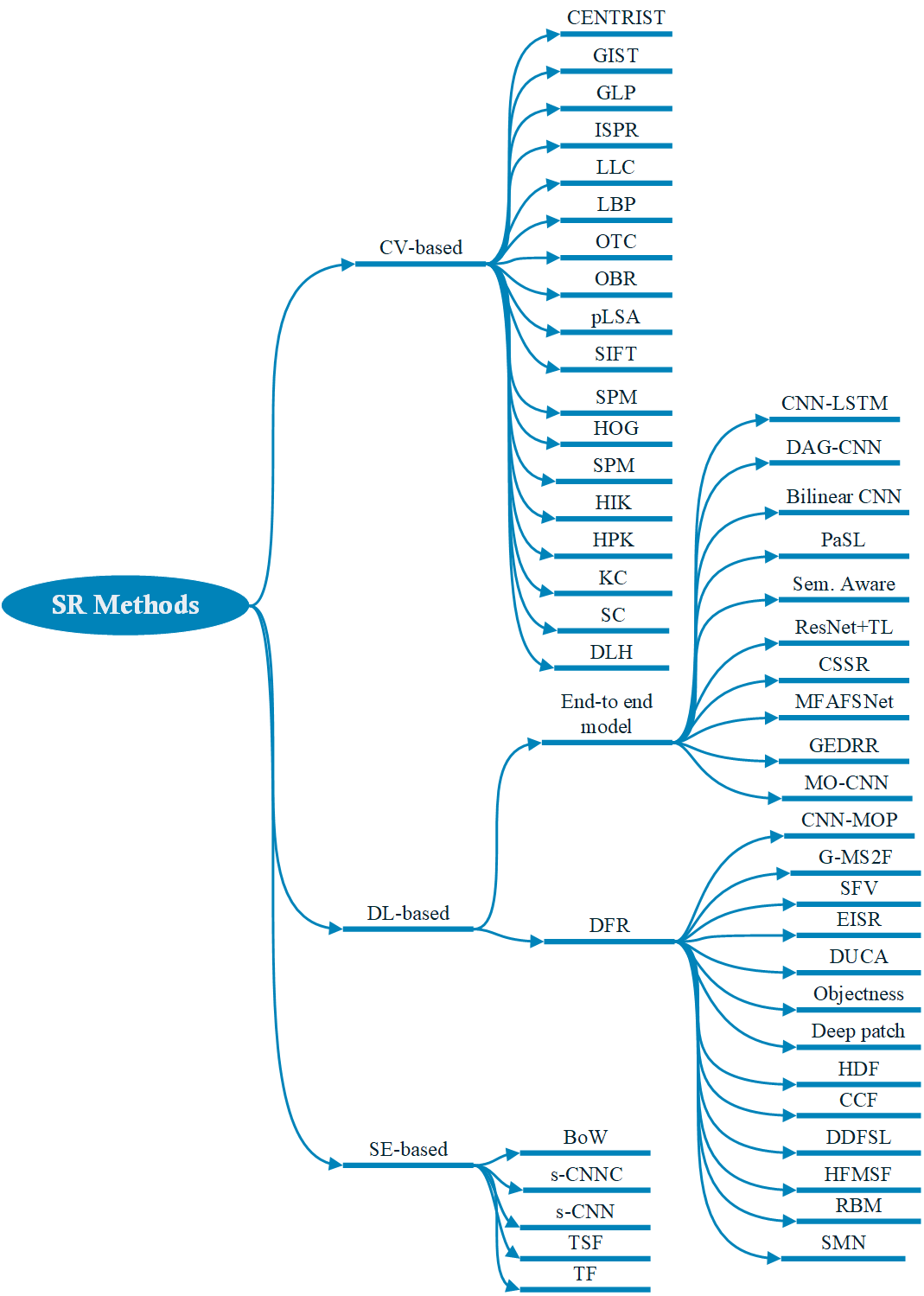}
    \caption{Taxonomy of existing scene image representation methods}
    \label{fig:texonomy}
\end{figure}

In this section, we categorize the existing scene representation methods into three broad categories, which are traditional CV-based, DL-based, and SE-based methods (refer to Fig. \ref{fig:texonomy} for the detailed taxonomy). The leaves of the taxonomy depict the algorithms for each method.
Each method is explained in detail in the next subsections. 

\subsection{Traditional computer vision (CV)-based methods}

\begin{figure}
    \centering
    \includegraphics[width=0.98\textwidth, height=100mm, keepaspectratio]{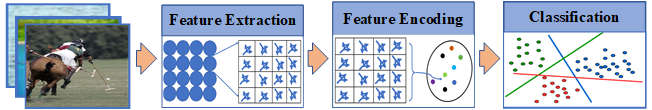}
    \caption{CV-based scene representation pipeline for classification}
    \label{fig:cv-based-pipeline}
\end{figure}

Traditional computer vision-based methods \cite{oliva_modeling_2001,zeglazi_sift_2016,wu_centrist:_2011, margolin2014otc, sinha2014new} are based on the basic components of the image such as colors, pixels, lines, and shapes. The use of such basic components helps us understand how images are constructed and based on such patterns, we can represent them easily for several tasks such as classification, clustering, recognition, and prediction. The high-level flow of traditional computer vision-based methods {for scene image representation and classification} is presented in Fig. \ref{fig:cv-based-pipeline}, which includes three steps: feature extraction, feature encoding, and classification.

Most popular traditional image representation methods are based on Generalized
Search Trees (Gist)~\cite{oliva2005Gist,oliva_modeling_2001}, Gist-Color ~\cite{oliva_modeling_2001}, CENsus TRansform hISTogram (CENTRIST)~\cite{wu_centrist:_2011},multi-channel (mCENTRIST)~\cite{xiao_mcentrist:_2014}, Scale-Invariant Feature Transform (SIFT) \cite{zeglazi_sift_2016}, Histogram of gradient(HoG)~\cite{dalal2005histograms}, Oriented Texture Curves (OTC)~\cite{margolin2014otc}, Object bank representation(OBR) ~\cite{li2010object,zhang2014learning}, SPM~\cite{lazebnik2006beyond}, Reconfigurable BoW (RBoW) ~\cite{parizi2012reconfigurable}, Bag of Parts (BoP) ~\cite{juneja2013blocks}, Important Spatial Pooling Region (ISPR) ~\cite{lin_learning_2014}, etc.
Among these techniques, the popular method such as Gist extracts the features from local details such as color, pixels, and orientation of images \cite{oliva_modeling_2001,quattoni_recognizing_2009,zhu_large_2010,li2010object,parizi2012reconfigurable,juneja2013blocks,lin_learning_2014,shenghuagao2010local,perronnin2010improving}. Therefore, they are limited to dealing with high variations in the local image features.
Furthermore, the OTC~\cite{margolin2014otc} method extracts the image features based on the color variation of various patches in images, keeping in mind that these features are suitable to represent the texture images, not much pertinent to scene images.
However, Spatial Pyramid Matching (SPM) \cite{lazebnik2006beyond} employs SIFT, which are multi-scale and rotation-invariant local features. Going forward, SPM first slices the images and then extract image feature based on those spatial regions of the image. The extracted features of each region are represented as a Bag of Visual Words (BoVW) of SIFT descriptors.
Even though this method captures more semantic regions than other methods of the scene image to some extent, they are still not suitable to represent complex scene images requiring high-level information such as object and foreground/background information for discriminability.


\subsection{Deep learning (DL)-based methods}
\begin{figure}
    \centering
    \includegraphics[width=0.98\textwidth, height=80mm, keepaspectratio]{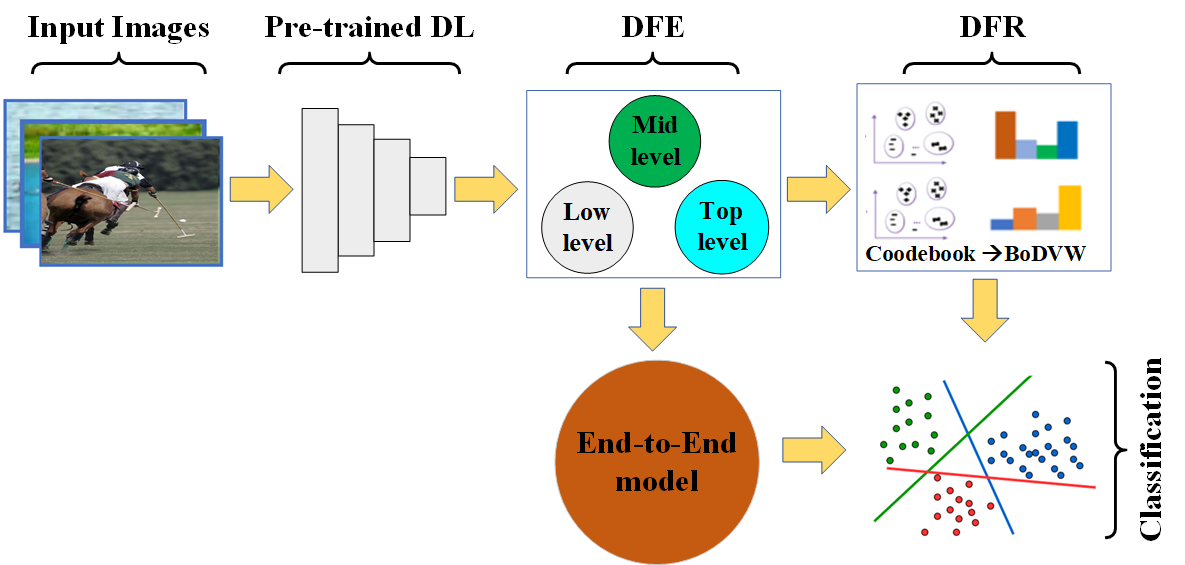}
    \caption{DL-based scene representation pipeline for classification}
    \label{fig:dl-based-pipeline}
\end{figure}

Deep learning models, which are a composition of multiple artificial neural networks \cite{lecun2015deep}, have provided a breakthrough performance in various domains such as text classification \cite{sitaula2021deep,shahi2021natural}, health informatics \cite{sitaula2022monkeypox} and computer vision \cite{sitaula2020hdf,shahi2022fruit}. Among three different methods, DL-based methods are most popular today to represent and classify scene images. The high-level diagram of DL-based methods is presented in Fig. \ref{fig:dl-based-pipeline}, which includes deep feature extraction (DFE) using pre-trained models (e.g., low-level, mid-level, and high-level), deep feature representation by encoding approach (e.g., a bag of words, fisher vector, etc.), and classification.
Besides, some DL methods prefer training in an end-to-end fashion after the deep feature extraction (DFE) step for the classification.

There are two approaches/techniques ({uni-modal and multi-modal}) preferred by most of the DL-based methods for scene image representation and classification. 
First, there are some works in scene representation and classification that use uni-modal pre-trained deep models such as ResNet152 \cite{he2016deep},  VGG-Net \cite{simonyan2014very, zhou2017places, bai2019coordinate}, AlexNet \cite{krizhevsky2012imagenet}, GoogleLeNet \cite{szegedy2015going}, and HDF  \cite{sitaula2020hdf}. 
For example, authors in \cite{gong_multi-scale_2014} extracted features from  VGG-Net pre-trained on hybrid datasets (ImageNet~\cite{deng_imagenet:_2009} and Places~\cite{zhou2016places}) using Caffe~\cite{jia2014caffe} platform. They used fully connected layers ($FC$), which resulted in a feature size of $4,096$-D for each scale of the image to achieve orderless multi-scale pooling features. The final feature size of their method is higher as the number of scales increases in their experiment. 
Their method outperforms the single-scaled features though their method has a higher dimensional feature size. 
Similarly, authors in \cite{kuzborskij2016naive} used features from VGG-Net pre-trained on ImageNet~\cite{deng_imagenet:_2009} and extracted the high-level feature from the $FC$-layers after a fine-tuning operation. These features were fed into the Naive Bayes non-linear algorithm \cite{fornoni2014scene} for the classification. The performance of their method is promising; however, their method requires a massive dataset for fine-tuning operations, which could limit its applicability in real time. 
Furthermore, authors in \cite{tang_g-ms2f:_2017} utilized three classification layers of fine-tuned GoogleNet \cite{szegedy2015going} model, where they extracted the deep features in the form of probabilities and then performed the features fusion to achieve the results. Although their method outperforms several existing methods in the literature, it requires large datasets for fine-tuning coupled with an arduous hyper-parameter tuning operation to learn the highly separable features.

Furthermore, some studies improved the separability of scene images by extracting the mid-level features from the pre-trained deep learning models. For instance, Zhang et al.~\cite{zhang2017image} randomly cropped the image into multiple patches and extracted the visual features from each of them using the AlexNet~\cite{krizhevsky2012imagenet} model. Then, these features were used to design the codebook of size $1,000$-D for the sparse coding technique to extract the relevant features. Later on, they concatenated the sparse coded features with the tag-based features to get the final features for the classification.
Because of highly discriminating features from both deep features and sparse coded features, their method imparts a significant boost in performance compared to the existing methods. However, their work possesses two main limitations: a) the chance of feature repetition as the patches are selected randomly; and b) higher feature size. 
In addition, bag of surrogate parts (BoSP) features were proposed by Guo et al. \cite{guo2016bag} based on the two higher pooling layers-- $4^{th}$ and $5^{th}$  of the VGG16 model \cite{simonyan2014very} pre-trained on ImageNet \cite{deng_imagenet:_2009}. 
However, their method only captures the foreground information as they employed the VGG-16 model pre-trained on ImageNet. As a result, it lacks the background information, which is one of the important clues required to better discriminate the complex scene images having higher inter-class similarity and intra-class dissimilarity. 
Additionally, authors in \cite{gupta2021recognition} compared four different CNN models such as AlexNet \cite{krizhevsky2012imagenet}, ResNet152 \cite{he2016deep}, VGG-16 \cite{simonyan2014very}, and GoogleLeNet \cite{szegedy2015going} pre-trained on ImageNet and Places datasets for scene image classification using semantic multinomial representation (SMN) approach, where they utilized pre-trained models available for Caffe \cite{jia2014caffe} model zoo without fully connected layers and fine-tuning operation. This is one of the recent methods used in scene image representation and classification, which has shown great promise against the existing methods.

\begin{table*}
\caption{Dataset description used in scene image representation and classification.}
\vspace{0mm}
\centering
\begin{tabular}{p{1.5cm} p{2cm} p{1.5cm} p{5cm} p{2cm}}
\toprule
Dataset& Type & Highlights & Ref.\\
\midrule
MIT-67&RGB & Complex scene images &
\cite{margolin2014otc,lin_learning_2014,xiao_mcentrist:_2014,zhang2013beyond,gong_multi-scale_2014,yang2015multi,tang_g-ms2f:_2017,dixit2015scene,zhou2016places,zhang2017image,wang_weakly_2017,guo_locally_2017,khan_discriminative_2016,cheng_scene_2018,7968351,bai2019coordinate,jiang2019deep,sitaula2020hdf,sorkhi2020comprehensive,chen2020scene,sitaula2020scene,sitaula2020content,lopez2020semantic,zhang2020locality,wang2019task,kim2014convolutional,sitaula2019tag,nascimento_robust_2017,sitaula2020scene,sitaula2020content}\\
\hline
 Scene-15&Grayscale & Indoor-outdoor images & 
 \cite{oliva_modeling_2001,lazebnik2006beyond,bosch2008scene,rasiwasia2008scene,van2009visual,wu_centrist:_2011,margolin2014otc,zhang2014object,lin_learning_2014,zhang2013beyond,yang2015multi,tang_g-ms2f:_2017,zhang2017image,khan_discriminative_2016,cheng_scene_2018,sitaula2020hdf,sorkhi2020comprehensive,sitaula2020content,kim2014convolutional,wang2019task,sitaula2019tag,xie2018improved,ali2018hybrid,nascimento_robust_2017,sitaula2020content}\\
\hline
 Event-8 &RGB & Sport events related images& 
 \cite{oliva_modeling_2001,wu_centrist:_2011,zhang2014object,lin_learning_2014,xiao_mcentrist:_2014,zhang2017image,khan_discriminative_2016,sitaula2020hdf,sorkhi2020comprehensive,sitaula2020content,wang2019learning,kim2014convolutional,sitaula2019tag,xie2018improved,sitaula2020content}\\
\hline
SUN-397 &RGB & Complex indoor/outdoor scene images &
\cite{margolin2014otc,gong_multi-scale_2014,yang2015multi,tang_g-ms2f:_2017,dixit2015scene,zhou2016places,wang_weakly_2017,guo_locally_2017,cheng_scene_2018,bai2019coordinate,jiang2019deep,chen2020scene,sitaula2020scene,lopez2020semantic,zhang2020locality,nascimento_robust_2017}\\
\hline
Caltech-256&RGB&Natural and artificial objects in a diverse setting&\cite{griffin2007caltech,van2009visual,wang2016csps,sinha2012novel}\\
\hline
NYU-V1&RGB-Depth&Indoor images with RGB and depth information&\cite{silberman2011indoor,ren2012rgb}\\


\bottomrule
\end{tabular}
\label{tab:dataset_description}
\end{table*}

Second, a few works proposed to use {multi-modal} deep features to represent the scene image for classification. For instance, Sun et al. \cite{sun2018fusing} used three models: YOLOV2 \cite{redmon2017yolo9000}, HybridDNN \cite{sun2018fusing}, and VGG-16 to represent the scene images. Here, the global appearance feature (GAF) from the second-last layer of VGG-16, CFA feature from the hybrid DNN and spatial layout maintained object semantics feature (SOSF) from the YOLOV2 models were concatenated to represent the scene image. The resultant features were trained using the SVM classifier.
Moreover, Bai et al. \cite{bai2019coordinate} proposed a multi-modal architecture utilizing both CNN and Long Short Term Memory (LSTM) model for the scene image classification. The LSTM model was used on top of CNNs. In their proposal, each image slice feature was extracted from VGG-16 \cite{simonyan2014very} pre-trained on Places \cite{zhou2016places} and then, fed into the LSTM model. Since the deep learning model pre-trained model on the Places dataset gives the background information and LSTM captures the sequence information of image slices, their model outperforms several other previous methods, including traditional CV-based methods and several DL-based methods. 
{Furthermore, Liu et al. \cite{liu2019indoor} proposed to use the CNN features and euclidean distance approach, which improved the performance on both MIT-67 and Scene-15 datasets.
 Furthermore, considering the popularity of metric learning and local manifold preservation, authors in \cite{liu2021scene} proposed a novel approach called, a joint global metric learning and local manifold preservation (JGML-LMP), which provided a significant boost in the classification performance. }

A few works on scene image classification used the whole-part feature extraction approach using both foreground and background information. For instance, the whole- and part-level feature extraction approach was proposed by  Sitaula et al. \cite{sitaula2020hdf} to represent the scene images. In their method, they utilized pre-trained VGG model on both ImageNet \cite{deng_imagenet:_2009} and Places \cite{zhou2016places} to capture both foreground and background information for each input scene image. Since their method does not consider contextual information, it still provides a limited performance while dealing with complex scene images having a higher inter-class similarity.{Authors in \cite{choe2021indoor} also employed the object-centric and place-centric information or features to classify the indoor images.}

\begin{figure}
    \centering
    \includegraphics[width=\textwidth, height=50mm, keepaspectratio]{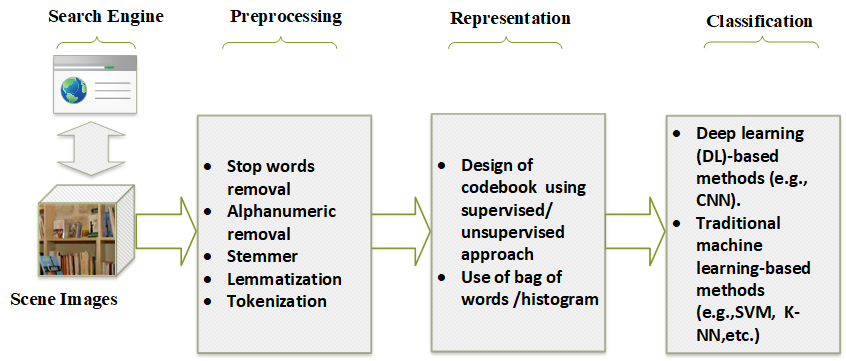}
    \caption{Search engine (SE)-based scene representation pipeline for classification.}
    \label{fig:se-based-pipeline}
\end{figure}

\subsection{Search engine (SE)-based methods}

The visual information achieved from either traditional CV-based or DL-based methods is not sufficient to represent the complex scene images because they also require contextual information (e.g., non-visual information such as tags, tokens, and annotation) for their accurate separability. 
There are very few works \cite{zhang2017image,wang2019task,sitaula2019tag}, which extract contextual information using a search engine, for the representation of scene images in the literature. These methods are considered SE-based methods.
While the extraction of features related to scene images using search engines is an arduous process, it still has an immense potential to differentiate complex scene images due to the presence of human annotations/descriptions for similar images on the web. The high-level diagram of SE-based methods is presented in Fig. \ref{fig:se-based-pipeline}, which comprises three steps:  preprocessing (e.g., stop words removal, stemmer, etc.), representation (e.g., codebook, histogram, etc.) and classification.

Under the SE-based methods, authors in \cite{zhang2017image} collated the annotations/tags of top $50$ visually similar searched images for the phrased input query image on the web. The collated tags were preprocessed and classified in an end-to-end fashion.
The main limitation of their work is the higher feature size incurred by the bag of words on raw tags, which could be minimized by using the filter bank.
Later on, the idea of filter banks to minimize the feature size was established by Wang et al.~\cite{wang2019task}, where they proposed the task-generic filter banks using the pre-defined category names to filter out the outlier tags to some extent. For the pre-defined category names, they borrowed them from  the ImageNet~\cite{deng_imagenet:_2009} and Places~\cite{zhou2016places} datasets.
However, their method still lacks domain-specific keywords/tags related to scene images, which could lead to out-of-vocabulary problems.
As a result, it creates an accumulation of unnecessary tags in the filter banks. This, in the end, could ultimately degrade the classification accuracy.
Given such limitations, Sitaula et al. \cite{sitaula2019tag} constructed the domain-specific filter bank based on the training data. Their domain-specific filter bank not only helped minimize the vocabulary problems but also improved the overall classification performance of scene images as they were able to capture more semantic information.
By and large, the contextual information captured from the web can provide important clues to discriminate complex scene images having both inter-class similarity and intra-class dissimilarity \cite{wang2019task,sitaula2019tag}.

\section{Datasets}
\label{datasets_performance}

Although several datasets, including both smaller and larger ones, have been used in the literature for scene representation and classification, we list and explain the commonly-used larger scene image datasets in this study. There are commonly six benchmark datasets (MIT-67 \cite{quattoni_recognizing_2009}, Scene-15 \cite{fei-fei_bayesian_2005}, Event-8 \cite{li2007and}, SUN-397 \cite{xiao2010sun}, Caltech-256 \cite{griffin2007caltech}, and NYU-V1 \cite{silberman2011indoor}), which have been used frequently in the literature.

{\bf MIT-67 \cite{quattoni_recognizing_2009}} contains $15,620$ images ($67$ categories), where each category contains at least $100$ images. There is a standard protocol \cite{quattoni_recognizing_2009} of train/test protocol to be used in the experiments. According to the protocol, $80$ images per category are taken as the training split, whereas $20$ images per category are taken as the testing split.

{\bf Scene-15 \cite{fei-fei_bayesian_2005}} contains $4,485$ images ($15$ categories), where each category contains at least $200$ images. There is no standard train/test protocol defined to use this dataset. However, researchers use $100$ images per category as training and the rest of the images as testing split. The experiment is repeated for $10$ runs to report the average accuracy.

{\bf Event-8 \cite{li2007and}} contains $1,579$ images ($8$ categories), where each category contains at least $137$ images. There is no standard train/test split ratio to use this dataset; however, researchers randomly select $120$ images per category and divide $70$ images as training and $60$ images per category as a testing split. The experiments are conducted for $10$ runs to note the average accuracy.  

{\bf SUN-397 \cite{xiao2010sun}} contains $108,754$ images ($397$ categories), where each category contains at least $100$ images. This dataset provides standard $10$ sets of train/test protocol \cite{xiao2010sun} to be used in the experiments, where each split contains $50$ images/category as training and $50$ images/category as testing. The average of $10$ runs is used to report the accuracy.

{\bf Caltech-256 \cite{griffin2007caltech}} contains $30,607$ images($256$ object categories). It consists of images of various natural and artificial objects in diverse settings. The minimum number of images in each category is 80.

{\bf NYU-V1 \cite{silberman2011indoor}} consists of $2,347$ labeled frames having 7 different classes. The images were collected from a wide range of domains, where the background was changing from one to another with RGB and depth cameras from the Microsoft Kinect. 
Given that scene images in this dataset contain several objects and their associations, this dataset is one of the most challenging datasets for scene image classification. 
Summary details of all of these datasets are mentioned in Table \ref{tab:dataset_description}.

\begin{table*}
\caption{Comparative study of state-of-the-art methods using classification accuracy (\%) on scene datasets under CV-based methods. The symbol $-$ represents the no published accuracy.}
\centering
\begin{tabular}{p{3cm}p{2cm} p{1.5cm} p{1.7cm} p{1.7cm}}
\toprule
 Approach&Scene-15 &Event-8 &MIT-67 &SUN-397\\
\midrule
Gist-color \cite{oliva_modeling_2001}&69.5 &70.7 &-&-\\
SPM \cite{lazebnik2006beyond}&72.2&- &- &- \\
pLSA \cite{bosch2008scene}&72.7&-&-&- \\
Semantic Theme \cite{rasiwasia2008scene}&72.2&-&- &- \\
Kernel Codebook \cite{van2009visual}&76.7&- &- &- \\
CENTRIST \cite{wu_centrist:_2011}&84.9&78.5&-&-\\
OTC \cite{margolin2014otc}&84.3&-&47.3&34.5\\
${S^3}R$ \cite{zhang2014object}&83.7&40.1&-&- \\
ISPR \cite{lin_learning_2014}&85.0&89.5&50.1&-\\ 
WSR-EC \cite{zhang2013beyond}&81.5&-&38.6&-\\
mCENTRIST \cite{xiao_mcentrist:_2014}&86.5&44.6&-&-\\
Xie et al. \cite{xie2018improved}&83.3&84.8&-\\
Ali et al. \cite{ali2018hybrid}&90.4&-&-\\
HIK\cite{niu2010hybrid} &-&-&40.19&-\\
HPK \cite{cho2012efficient}&-&-&-&-\\

HPK \cite{cho2012efficient}&-&-&-&-\\
HILLC \cite{chen2018scene-hillc}&86.3&85.0&-&-\\
CS-PSL \cite{wang2016csps}&-&-&52.5&-\\
OBR \cite{zhang2014learning}&88.8&86.0&32.3&-\\
3-DLH \cite{banerji2012novel}&-&84.9&-&-\\
LLC \cite{li2012reference}&83.2&-&-&-\\
PFE \cite{li2013representative}&84.2&-&-&-\\
SIFT\cite{silberman2011indoor}&-&-&-&-\\
W-LBP\cite{sinha2014scene}&85.1&86.2&-&-\\
GPHOG \cite{sinha2014new}&-&-&-&-\\
Spatial LBP \cite{hu2012spatial}&80.9&71.7&-&-\\
BoW-LBP \cite{banerji2012novel}&80.7&87.7&-&-\\
\bottomrule
\end{tabular}
\label{tab:comparision_study11}
\end{table*}

\begin{table*}
\caption{Comparative study of state-of-the-art methods using classification accuracy (\%) on four scene datasets under DL-based methods. The symbol $-$ represents the no published accuracy.}
\centering
\begin{tabular}{p{3cm}p{2cm} p{1.5cm} p{1.7cm} p{1.7cm}} 
\toprule
 Approach&Scene-15 &Event-8 &MIT-67 &SUN-397\\ 
\midrule
CNN-MOP \cite{gong_multi-scale_2014}&-&-&68.8&51.9\\
DAG-CNN \cite{yang2015multi}&92.9&-&77.5&56.2\\
G-MS2F \cite{tang_g-ms2f:_2017}&92.9 &- &79.6&64.0\\
SFV+Places \cite{dixit2015scene}&-&-&79.0&61.7\\
VGG \cite{zhou2016places}&91.72&95.17 &79.7&63.2 \\
EISR \cite{zhang2017image}&92.1&89.6&66.2&- \\
VSAD \cite{wang_weakly_2017}&-&-&86.2&73.0\\
LS-DHM \cite{guo_locally_2017}&-&-&83.7&67.5\\
DUCA \cite{khan_discriminative_2016} &94.5 &98.7 &71.8 &-\\
Nascimento et al. \cite{nascimento_robust_2017}&95.7&-&87.2&71.0\\
Objectness \cite{cheng_scene_2018}&95.8&-&86.7&73.4\\
Bilinear-CNN \cite{7968351}&-&-&79.0&-\\
Deep patch \cite{jiang2019deep}&-&-&79.6&57.4\\
HDF \cite{sitaula2020hdf}&93.9&96.2&82.0&-\\
Sorkhi et al. \cite{sorkhi2020comprehensive}&95.1&99.2&73.6&-\\
PaSL \cite{chen2020scene}&-&-&88.0&74.0\\
Semantic-Aware \cite{lopez2020semantic}&-&-&87.1&74.0\\
LASC \cite{zhang2020locality}&-&-&81.7&64.3\\
FBH \cite{sitaula2020scene}&-&-&82.3&66.3\\
CCF \cite{sitaula2020content}&95.4&98.1&87.3&-\\
DDSFL \cite{zuo2015exemplar}&52.2&86.9&84.4&-\\

ResNet+TL\cite{liu2019novel}&85.2&-&94.0&-\\
HFMSF\cite{khan2021image}&97.8&-&-&-\\
CNN-LSTM\cite{bai2019coordinate}&-&-&80.5&63.0\\
ABR \cite{liu2018attribute}&91.9&96.2&68.3&-\\
CSSR \cite{qi2016cssr}&-&-&77.8&57.3\\
RBM \cite{xie2018class}&98.7&-&-&-\\
SOSF+CFA+GAF \cite{sun2018fusing} &-&-&89.5&78.9\\
DeepFeature \cite{bai2017growing} &-&94.8&72.3&-\\
SMN \cite{gupta2021recognition} &-&-&84.4&66.8\\
RVF \cite{sharma2018scene}&-&-&80.0&60.6\\
MFAFSNet \cite{dixit2019semantic}&-&-&88.0&72.4\\
GEDRR \cite{lin2021global}&96.0&-&87.7&73.5\\
MetaObject +CNN \cite{wu2015harvesting}&-&-&78.9&58.1\\
{JGML-LMP\cite{wang2022joint}} &96.0&99.0&87.5&73.2\\
{Liu et al. \cite{liu2021scene}} &96.4&-&81.6&-\\
{Selective CNN \cite{liu2021scene}} &-&-&88.4&-\\
\bottomrule
\end{tabular}
\label{tab:comparision_study21}
\end{table*}

\begin{table*}
\caption{Comparative study of state-of-the-art methods using classification accuracy (\%) on four scene datasets under SE-based methods. The symbol $-$ represents the no published accuracy.}
\centering
\begin{tabular}{p{3cm}p{1.7cm} p{1.5cm} p{1.5cm} p{1.5cm}}
\toprule
Approach&Scene-15 &Event-8 &MIT-67\\
\midrule
BOW \cite{wang2019task}&70.1 &83.5 &52.5&\\
s-CNN(max) \cite{wang2019task} &76.2&90.9&54.6& \\
s-CNN(avg) \cite{wang2019task}&76.7&91.2&55.1&\\
s-CNNC(max) \cite{wang2019task}&77.2&91.5&55.9&\\
TSF \cite{sitaula2019tag}&81.3&94.4&76.5&\\
TF \cite{sitaula2020content}&84.9&95.8&77.1&\\
\bottomrule
\end{tabular}
\label{tab:comparision_study3}
\end{table*}

\section{Discussion}
\label{discussion}
Here, we discuss the research works carried out in scene representation and classification using quantitative (e.g., performance metrics) and qualitative analysis (e.g., pros/cons). 

\subsection{Quantitative analysis}

\begin{figure*}
\begin{center}
 \subfloat[]{\includegraphics[width=65mm, height=40mm,keepaspectratio]{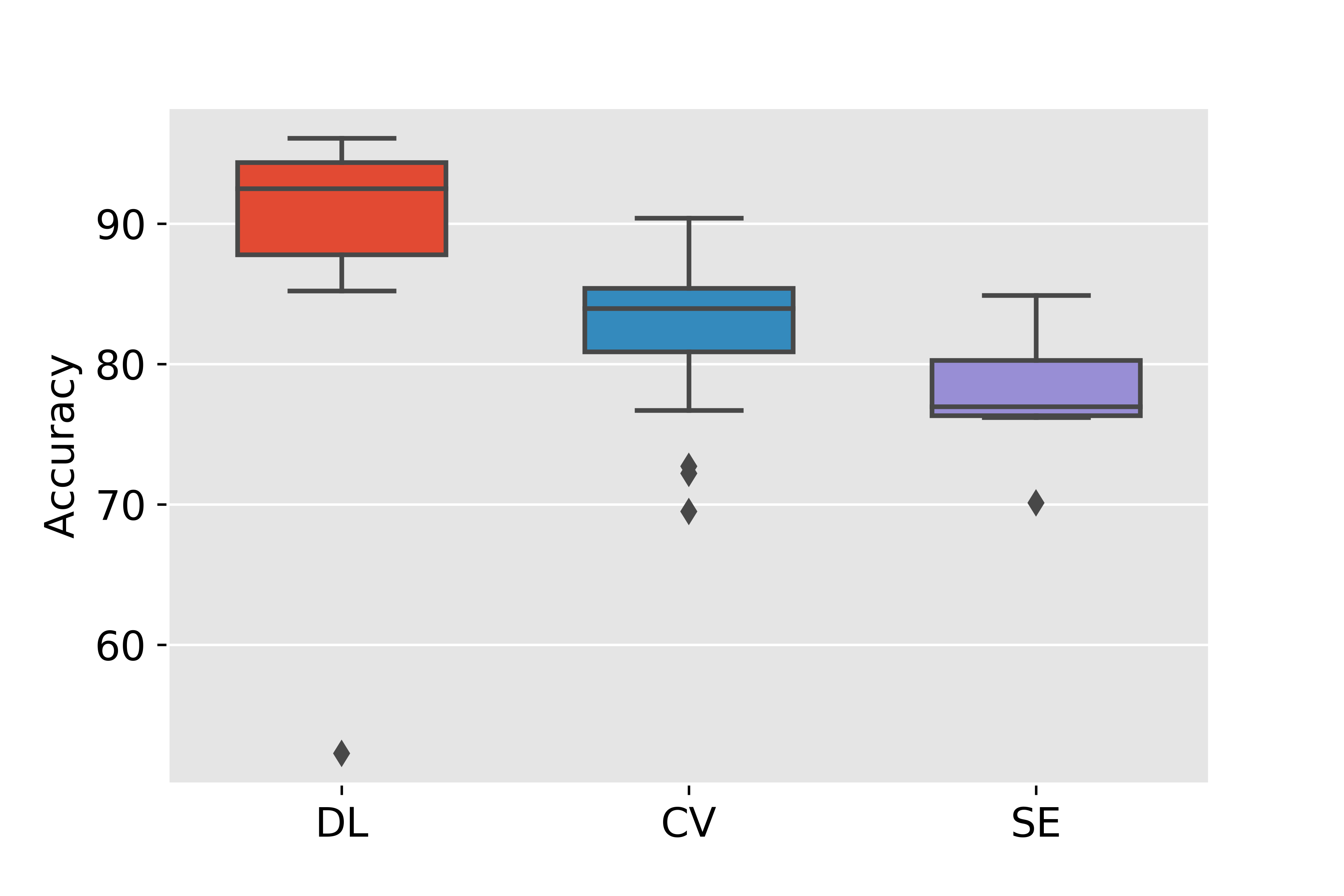}}
  \hspace{0pt}
 \subfloat[]{\includegraphics[width=65mm, height=40mm,keepaspectratio]{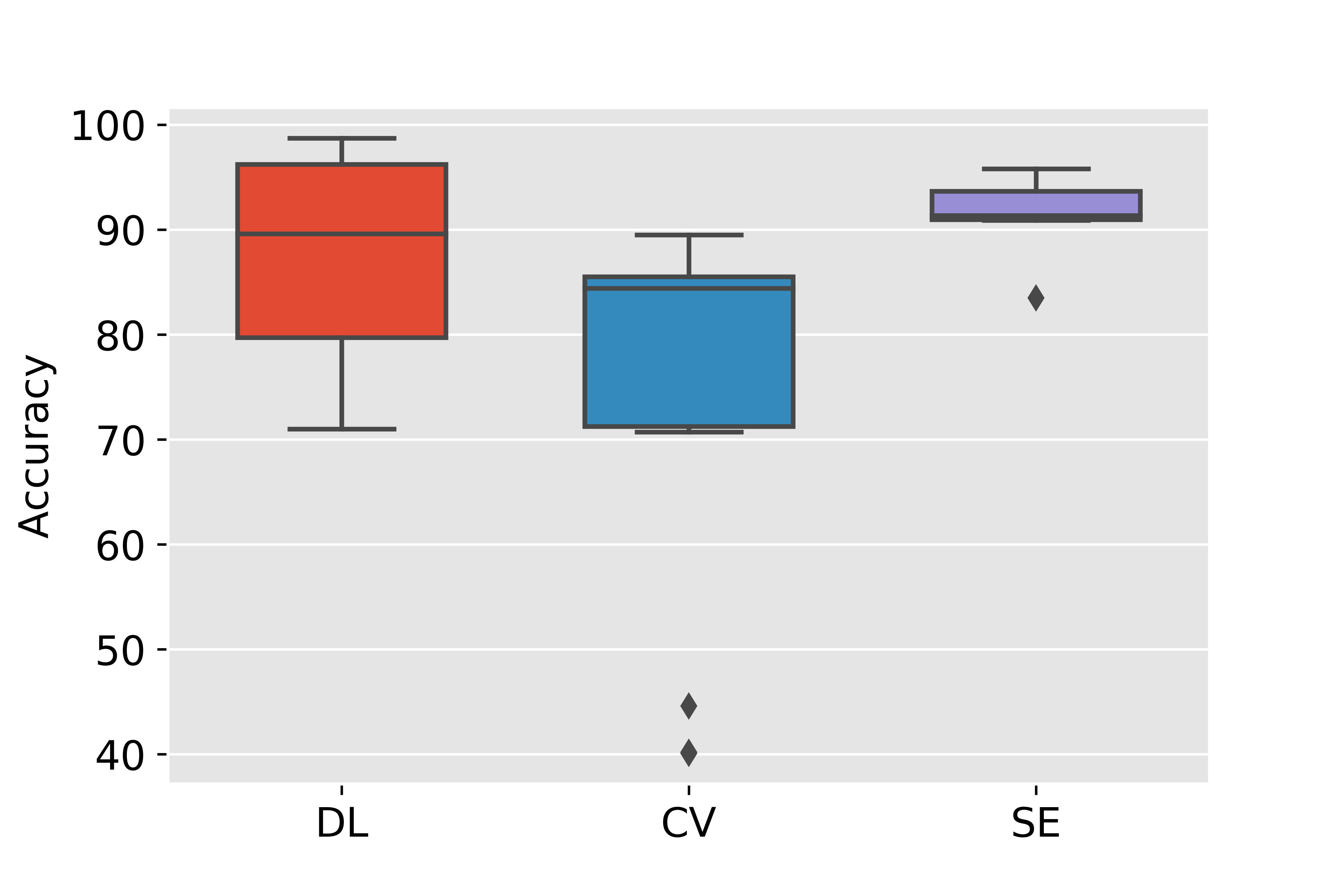}}
 \hspace{0pt}
 \subfloat[]{\includegraphics[width=65mm, height=40mm,keepaspectratio]{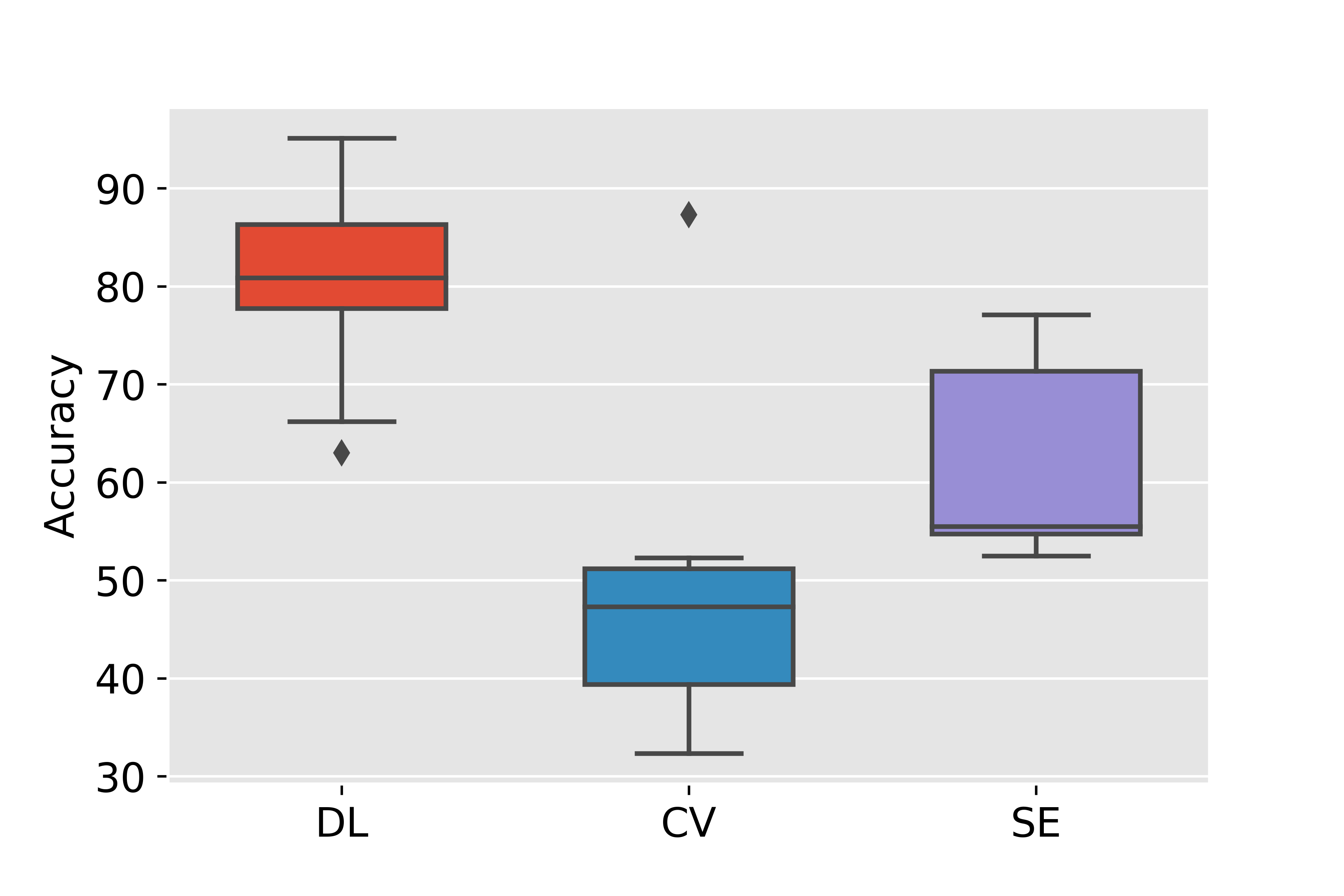}}
 \hspace{2pt}
  \subfloat[]{\includegraphics[width=65mm, height=40mm,keepaspectratio]{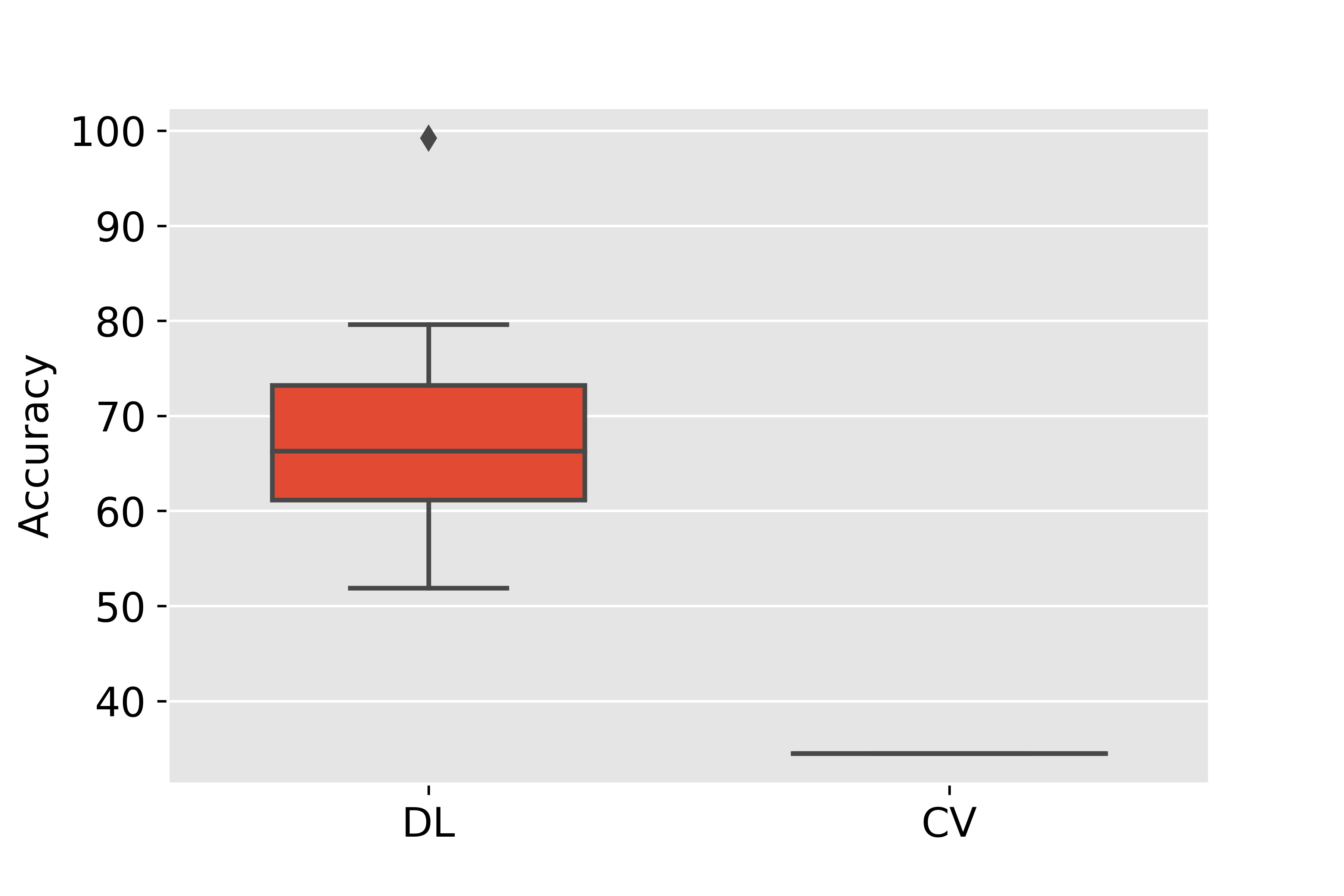}}
  \hspace{2pt}
  \caption{Box-plot visualization of summary accuracy (\%) achieved by three different methods on four most popular scene image datasets: (a) Scene-15, (b) Event-8, (c) MIT-67, and (d) SUN-397. 
   Note that DL, CV, and SE represent DL-based, CV-based, and SE-based methods. Note that there is no reported accuracy for SE-based methods on the SUN-397 dataset.
  }
  \label{fig:box_plot_analysis}
 \end{center}
 \end{figure*}

For the quantitative analysis of research articles published in the literature, we summarize the performance using box plots, which impart the statistical information of classification performance, as shown in Fig. \ref{fig:box_plot_analysis}.
(Note that we draw boxplots based on the performance of three different scene representation methods (DL-based, CV-based and SE-based ) achieved from the corresponding Tables \ref{tab:comparision_study11}, \ref{tab:comparision_study21} and \ref{tab:comparision_study3} on four datasets  (Figs. \ref{fig:box_plot_analysis}(a), \ref{fig:box_plot_analysis}(b),  \ref{fig:box_plot_analysis}(c) and
\ref{fig:box_plot_analysis} (d),  respectively.)

Here, we analyze the performance, particularly the reported accuracies of three or two different methods on four datasets. Since the search engine (SE)-based methods only consider three datasets (Scene-15, Event-8, and MIT-67) in the literature, we present the results on only such three datasets, whereas, for the other two methods (DL-based and CV-based), we present the results on four datasets (Scene-15, Event-8, MIT-67, and SUN-397).

While comparing the performance of three different kinds of methods on four datasets, we notice that DL-based methods outperform other remaining methods in all datasets. For example, on the Scene-15 dataset, DL-based methods provide the highest accuracy (maximum and minimum of over 98\%, and over 85\%, respectively) compared to the traditional CV-based methods (below 85\%). The reason for such performance surge while using DL-based methods is because of the highly discriminating feature extraction abilities from different intermediate layers of DL methods. Notably, deep features could provide more information related to scene images, including foreground, background, and hybrid. The presence of all three kinds of information helps discriminate the complex scene images more accurately. However, traditional CV-based methods are not sufficient to capture such information, which as a result fails to discriminate the complex scene images during classification. Also, the recent works using the search engine (SE)-based methods on three datasets (Scene-15, Event-8, and MIT-67) show that SE-based methods could capture complementary contextual information, which is difficult to achieve from the visual information achieved from the traditional CV-based and DL-based methods, for the scene images to represent them during classification. Interestingly, it can outperform the traditional CV-based methods and is comparable to DL-based methods during scene image representation and classification. For example, SE-based methods on the Event-8 dataset (\ref{fig:box_plot_analysis}(b)) provide an accuracy of over 90\%, whereas the traditional CV-based methods and DL-based methods provide an accuracy below 90\% and over 90\%, respectively. This encouraging classification performance shows the efficacy of SE-based methods for scene image representation.

While comparing the performance throughout the four widely popular datasets (Scene-15, Event-8, MIT-67, and SUN-397) reported in Fig. \ref{fig:box_plot_analysis}, we observe that SUN-397 is the most challenging dataset for which the state-of-the-art methods have produced the least performance compared to the other three datasets (Scene-15, Event-8, and MIT-67). Also, there is no reported classification accuracy for SE-based methods for this dataset. Furthermore, the accuracy of SUN-397 remains between around 71\% and 35\% in the classification. We believe that this is the most challenging dataset compared to other datasets, both in terms of complexities (higher inter-class similarity and intra-class dissimilarity) and categories (higher number of challenging classes). Similarly, we observe that the MIT-67 dataset is the second-most challenging dataset in terms of performance, which has a maximum performance of around {97\%} by DL-based methods and a minimum performance of around 40\% by CV-based methods. Although this dataset has only 67 categories compared to SUN-397 (397 categories), it is still a challenging dataset with a similar level of complexity to SUN-397 for scene image representation and classification.  Compared to the SUN-397 and MIT-67 datasets, two other datasets (Scene-15 and Event-8) are relatively less challenging and have produced the most prominent classification performance (Scene-15 has the maximum and minimum accuracy of over 98\% by DL-based methods and over 76\%, by SE-based methods respectively, whereas the Event-8 has the maximum and minimum accuracy of over 95\% by DL-based methods and over 70\% by CV-based methods, respectively). The reason for such a significant boost in performance is attributed to the distinguishable scene images (lower inter-class similarity and intra-class dissimilarity) present in them.

To sum up, the DL-based methods outperform both the traditional CV-based method and SE-based methods in most cases. This infers that visual content information of the scene images provided by the DL-based methods is more discriminating than others to distinguish ambiguous and complex scene images. Recently, the SE-based methods have shown some promise in scene image representation by providing some important contextual clues, which are attained using human perception and knowledge available on the internet.

\subsection{Qualitative analysis}
Here, we analyze each of the three methods (CV-based, DL-based, and SE-based) based on their advantages and shortcomings, which are obtained in terms of their viability.

Regarding CV-based methods, they have four major merits. First, feature extraction is well-established and easier to implement. For example, we can achieve the features based on the traditional CV-based methods such as SIFT (Scale Invariant Feature Transform) and HoG (Histogram of Gradient) with a few lines of code. Second, they have a higher performance with fine-grained and non-ambiguous images (no inter-class similarity and intra-class dissimilarity). With the help of basic information of scene images such as pixels, lines, and arc details, it is easy to distinguish the non-complex images (e.g., fine-grained, texture, non-ambiguous, etc.) during classification. Third, CV-based methods are less complex compared to other methods because they do not require arduous training activities to achieve the discriminating features of the input image.
Fourth, we do not require {a domain-specific knowledge} to implement them. {For example, we can apply the same SIFT algorithm for both scene images and biomedical images to represent them.}  
In contrast, CV-based methods have two major demerits. First, they have a lower classification performance for complex scene images having higher inter-class similarity and intra-class dissimilarity. This is because complex scene images require a higher level of information (e.g., object), which is difficult to acquire by CV-based methods. Second, given that there are several kinds of features achieved from the CV-based methods, it is very difficult to choose the most discriminating and useful features corresponding to the study.

For the DL-based methods, they have two major merits. First, they have a higher classification performance on complex images compared to CV-based methods. This is because they can extract the high-level information (e.g., object) present in the scene image. Second, DL-based methods are flexible. That is, the DL models can be re-trained using custom datasets unlike the CV-based methods to make them domain-specific.
Nevertheless, DL-based methods have three major demerits. First, they are heavy-weight in most cases compared to CV-based methods. The DL-based methods are very difficult to deploy in the edge computing environment as they require heavily trained weight files to achieve promising accuracy. Second, the training and re-training processes of DL-based models are labor-intensive as they are prone to over-fitting and under-fitting problems. Third, although they have higher accuracy compared to others, they are, in most cases, poor in interpretability and explainability.

The SE-based methods have two major merits. First, they can capture contextual information with the help of human knowledge, which is complementary information to visual features. Second, the combination of contextual information with visual information could overcome the limitations of each individual.
In contrast, they have two major demerits. First, they are computationally infeasible to capture the information via search engines if we have a massive number of images because search engines have a restriction on the number of query inputs for searching. Second, while selecting the tokens or textual information online, it is very difficult to select the most important information from the annotations/tags as we encounter numerous significant pieces of information. Since the current works focus on top-k images for annotations/tags, they could end up missing some important information present beyond k images.

\subsection{Research trend analysis}

\begin{figure*}[tb]
    \begin{center}
     \subfloat[]{\includegraphics[width=0.85\textwidth, height=90mm, keepaspectratio]{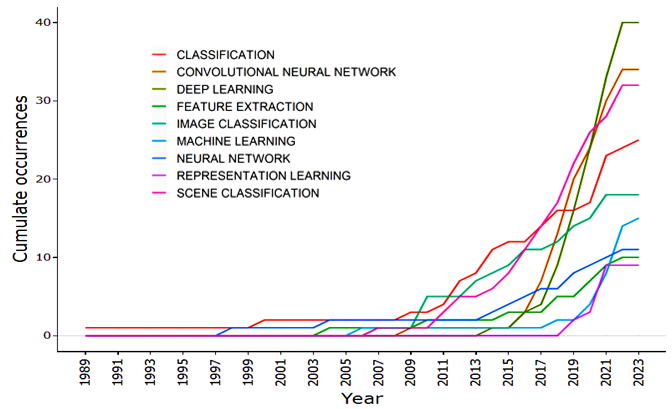}}
     \hspace{1pt}
     \subfloat[]{\includegraphics[width=0.85\textwidth, height=85mm, keepaspectratio]{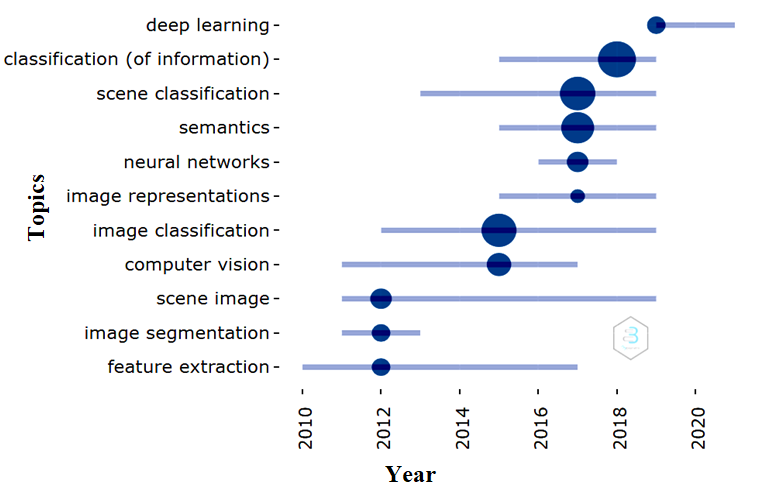}}
    \caption{Author's keyword growth during last decades}
    \label{fig:word_dynamics}
    \end{center}
\end{figure*}

Here, we analyze the research direction of scene representation based on the cumulative occurrence of keywords and time duration across different years using a Line graph and Forest plot \cite{aria2017bibliometrix}, respectively, which are presented in Fig. \ref{fig:word_dynamics}. {The frequently-used keywords help understand the research direction in scene analytics because they not only provide the frequency but also their inception and current state. In this study, such keywords have been picked by the Forest plot automatically based on their importance.}

While looking at Fig. \ref{fig:word_dynamics} in terms of topic occurrence, we observe that the cumulative topic occurrence has been increasing from 1996 to this date. There have been several topics popular in scene image representation such as 'classification (of information)', 'computer vision, 'deep learning, and 'semantic'. Among them, it is noted that 'classification(of information)' is the most popular topic, which has been sharply increasing in recent years. In addition, some other topics such as 'scene classification', and 'feature extraction' are also following similar kinds of patterns, whereas other topics such as image segmentation and scene classification are increasing at a slower rate. We believe that this trend makes sense because basic works related to scene image representation have already been done such as 'scene classification' and 'feature extraction. The current need is to build robust AI models with higher performance. Overall, the research trend of different topics in scene images has been in the upward direction with the predominant use of DL-based methods.  

While analyzing the keyword topics' popularity in terms of time duration at Fig. \ref{fig:word_dynamics}, we notice that different topics have different time duration for their popularity level. For example, from 2010 to 2017, most of the research works in scene representation were focused on feature extraction and it was most popular in 2012. We believe that this is because feature extraction is the foundation work of scene image representation.  
It is seen that most of the research topics in scene image representation such as 'semantics', 'neural networks, 'scene classification', and 'classification' are quite popular after 2017. In recent days, particularly after 2019, 'deep learning has become a prominent topic, which is because of the groundbreaking classification performance produced by them. To this end, the popularity of different keywords in different years reveals the different levels of research in scene representation and classification.

\section{Conclusion}
\label{conclusion}
In this paper, we have reviewed the research works carried out in the scene image representation area for classification and categorized them into three broad groups: 
 methods, DL-based methods, and SE-based methods. This categorization and analysis (both qualitative and quantitative) reveals that DL-based methods outperform the remaining two methods in terms of classification accuracy in most cases, whereas SE-based methods remain the potential research direction in the future. Through this study, we underline that the combination or fusion of DL-based methods with other methods enhances the classification performance significantly, which is because of the rich information obtained from multiple sources during image representation. {In addition, we find that scene representation research is on the rise in recent years.}

Furthermore, we notice that the usability of the method for the scene image representation is dependent on our requirements. If the requirement is on a performance issue, it is inevitable to use the DL-based methods as they have a groundbreaking performance track; however, they require higher computational and space requirements. To deal with it, we encourage building the domain-specific lightweight pre-trained DL model to be used in the future. {Also, the SE-based methods could also be interesting to capture the complementary information for more accurate representation during classification.}

\section{Data availability}
All data are publicly available.

\section{Abbreviations}
The list of abbreviations used in our study is presented in Table \ref{tab:list_of_abbreviations}. 
\begin{table}
    \centering
   
    \begin{tabular}{p{1.5cm}|p{9.5cm}}
    \toprule
    \hline
    Abbrv. & Full form\\
    \midrule
         ABR & Attribute-Based high-level image Representation \\
         BSRC & Block Sparse Representation Based Classifier\\
         CCF & Content Context Features \\
         CFA & Contextual Features in Appearance\\
         CSSR &Category-Specific Salient Region \\
         CS-PSL &class-specific pooling shapes Learning  \\
         DDSFL &Deep Discriminative and Shareable Feature Learning  \\
         DoG & Difference of Gaussian\\
         DAG-CNN &Directed Acyclic graph-Convolution Neural Network \\
         DUCA & Deep Un-structured Convolutional Activation \\
         EISR & Explicitly and Implicitly Semantic Representations  \\
         FBH & Foreground background hybrid features \\
         GAF & Global Appearance Feature \\
         GEDRR&Global and Graph Encoded Local Discriminative Region Representation \\
         Gist & Generalized Search Trees\\
         GPHOG & Gabor Pyramid of Histograms of Oriented Gradients  \\
         G-MS2F &GoogLeNet-based Multi-Stage Feature Fusion  \\
         GMM & Gaussian Mixture Model\\
         HDF & Hybrid deep features \\
         HFMSF & Handcrafted Features with Deep Multi-stage Features  \\
         HIK & Histogram Intersection Kernel\\
         HILLC &Histogram Intersection-Locally-constrained Linear coding  \\
         HPK & Hybrid Pyramid Kernel \\
         ISPR &Important Spatial Pooling Region  \\
         IoT& Internet of Things \\
         LoG &Laplacian of Gradient\\
         LASC & Locality-constrained Affine Subspace Coding  \\
         LS-DHM & Locally Supervised Deep Hybrid Model  \\
         LSTM& Long short-term memory \\
         MFAFSNet& Mixture of Factor Analyzers-Fisher Score Network \\
         MOP &Multiscale orderless pooling  \\
         OTC & Oriented Texture Curves  \\
         OBR &  Object Based Representation\\ 
         pLSA & probabilistic Latent Semantic Analysis \\
         PFE & Pooled Feature Extraction \\
         RBM & Restricted Boltzman Machine \\
         RVF &Reduced Virtual Features \\
         SC & Sparse coding\\
         SIFT& Scale-Invariant Feature Transform \\
         SOSF& Spatial-layout maintained Object Semantics Features \\
         SPM & Spatial Pyramid Matching \\
         SMN &semantic Multinomial Network  \\
         ${S^3}R$ &Sub-semantic space\\
         SFV&Semantic Fisher Vectors  \\
         TSF &Tag-based semantic features \\
         TF &Tag-based features\\
         VGG & Visual Geometry Group \\
         VSAD  & Vector of Semantically Aggregating Descriptor  \\
         W-LBP &Wigner-based Local Binary Patterns   \\
         WSR-EC & Weak semantic image representation- Example classifier \\
         3-DLH & 3-Dimensional LBP-HaarHOG \\
         \hline
    \bottomrule
    \end{tabular}
     \caption{List of abbreviations used in this study}
    \label{tab:list_of_abbreviations}
\end{table}

\clearpage

\section{Conflict of Interest}
The authors declare that they have no conflict of interest.

\bibliographystyle{spbasic}

\bibliography{sample}

\begin{thebibliography}{118}
\providecommand{\natexlab}[1]{#1}
\providecommand{\url}[1]{{#1}}
\providecommand{\urlprefix}{URL }
\expandafter\ifx\csname urlstyle\endcsname\relax
  \providecommand{\doi}[1]{DOI~\discretionary{}{}{}#1}\else
  \providecommand{\doi}{DOI~\discretionary{}{}{}\begingroup
  \urlstyle{rm}\Url}\fi
\providecommand{\eprint}[2][]{\url{#2}}

\bibitem[{Sitaula et~al(2019)Sitaula, Xiang, Zhang, Lu, and
  Aryal}]{sitaula2019indoor}
Sitaula C, Xiang Y, Zhang Y, Lu X, Aryal S (2019) {Indoor image representation
  by high-level semantic features}. IEEE Access 7:84,967--84,979

\bibitem[{Shadman~Roodposhti et~al(2019)Shadman~Roodposhti, Aryal, Lucieer, and
  Bryan}]{shadman2019uncertainty}
Shadman~Roodposhti M, Aryal J, Lucieer A, Bryan BA (2019) Uncertainty
  assessment of hyperspectral image classification: Deep learning vs. random
  forest. Entropy 21(1):78

\bibitem[{Neupane et~al(2021)Neupane, Horanont, and Aryal}]{neupane2021deep}
Neupane B, Horanont T, Aryal J (2021) Deep learning-based semantic segmentation
  of urban features in satellite images: A review and meta-analysis. Remote
  Sensing 13(4):808

\bibitem[{Dutta et~al(2013)Dutta, Aryal, Das, and Kirkpatrick}]{dutta2013deep}
Dutta R, Aryal J, Das A, Kirkpatrick JB (2013) Deep cognitive imaging systems
  enable estimation of continental-scale fire incidence from climate data.
  Scientific reports 3(1):1--4

\bibitem[{Sitaula et~al(2019)Sitaula, Xiang, Aryal, and
  Lu}]{sitaula2019unsupervised}
Sitaula C, Xiang Y, Aryal S, Lu X (2019) {Unsupervised deep features for
  privacy image classification}. In: Proc. Pacific-Rim Symposium on Image and
  Video Technology (PSIVT), pp 404--415

\bibitem[{McCulloch and Pitts(1943)}]{mcculloch1943logical}
McCulloch WS, Pitts W (1943) A logical calculus of the ideas immanent in
  nervous activity. The bulletin of mathematical biophysics 5(4):115--133

\bibitem[{Sitaula et~al(2019)Sitaula, Xiang, Basnet, Aryal, and
  Lu}]{sitaula2019tag}
Sitaula C, Xiang Y, Basnet A, Aryal S, Lu X (2019) {Tag-based semantic features
  for scene image classification}. In: Proc. Int. Conf. on Neural Inf. Process.
  (ICONIP), pp 90--102

\bibitem[{Wei et~al(2016)Wei, Phung, and Bouzerdoum}]{wei2016visual}
Wei X, Phung SL, Bouzerdoum A (2016) Visual descriptors for scene
  categorization: experimental evaluation. Artif Intell Rev 45(3):333--368

\bibitem[{Anu and Anu(2016)}]{anu2016survey}
Anu E, Anu K (2016) A survey on scene recognition. Int J Sci Eng Technol
  Res(IJSETR) 5:64--68

\bibitem[{Singh et~al(2017)Singh, Girish, and Ralescu}]{singh2017image}
Singh V, Girish D, Ralescu A (2017) Image understanding-a brief review of scene
  classification and recognition. In: Proc. Modern Artificial Intelligence and
  Cognitive Science (MAICS), pp 85--91

\bibitem[{Xie et~al(2020)Xie, Lee, Liu, Kotani, and Chen}]{xie2020scene}
Xie L, Lee F, Liu L, Kotani K, Chen Q (2020) Scene recognition: a comprehensive
  survey. Pattern Recognit p 107205

\bibitem[{Lowe(1999)}]{lowe1999object}
Lowe DG (1999) Object recognition from local scale-invariant features. In:
  Proc. Int. Conf. Comput. Vis. (ICCV), vol~2, pp 1150--1157

\bibitem[{Lazebnik et~al(2006)Lazebnik, Schmid, and Ponce}]{lazebnik2006beyond}
Lazebnik S, Schmid C, Ponce J (2006) {Beyond bags of features: Spatial pyramid
  matching for recognizing natural scene categories}. In: Proc. IEEE Comput.
  Soc. Conf. Comput. Vis. Pattern Recognit. (CVPR), pp 2169--2178

\bibitem[{Moller et~al(1997)Moller, Machiraju, Mueller, and
  Yagel}]{moller1997evaluation}
Moller T, Machiraju R, Mueller K, Yagel R (1997) Evaluation and design of
  filters using a taylor series expansion. IEEE transactions on Visualization
  and Computer Graphics 3(2):184--199

\bibitem[{Dalal and Triggs(2005)}]{dalal2005histograms}
Dalal N, Triggs B (2005) {Histograms of oriented gradients for human
  detection}. In: Proc. IEEE Comput. Soc. Conf. Comput. Vis. Pattern Recognit.
  (CVPR), pp 886--893

\bibitem[{Xie et~al(2018)Xie, Lee, Liu, Yin, Yan, Wang, Zhao, and
  Chen}]{xie2018improved}
Xie L, Lee F, Liu L, Yin Z, Yan Y, Wang W, Zhao J, Chen Q (2018) Improved
  spatial pyramid matching for scene recognition. Pattern Recognition
  82:118--129

\bibitem[{Zabih and Woodfill(1994)}]{zabih1994non}
Zabih R, Woodfill J (1994) Non-parametric local transforms for computing visual
  correspondence. In: Proc. Euro. Conf. Comput. Vis. (ECCV), pp 151--158

\bibitem[{Xiao et~al(2014)Xiao, Wu, and Yuan}]{xiao_mcentrist:_2014}
Xiao Y, Wu J, Yuan J (2014) mcentrist: a multi-channel feature generation
  mechanism for scene categorization. IEEE Trans Image Process 23(2):823--836

\bibitem[{Margolin et~al(2014)Margolin, Zelnik-Manor, and
  Tal}]{margolin2014otc}
Margolin R, Zelnik-Manor L, Tal A (2014) {OTC: A novel local descriptor for
  scene classification}. In: Proc. Eur. Conf. Comput. Vis. (ECCV), pp 377--391

\bibitem[{Sitaula et~al(2021{\natexlab{a}})Sitaula, Xiang, Aryal, and
  Lu}]{sitaula2020scene}
Sitaula C, Xiang Y, Aryal S, Lu X (2021{\natexlab{a}}) Scene image
  representation by foreground, background and hybrid features. Expert Systems
  with Applications p 115285

\bibitem[{Sitaula et~al(2021{\natexlab{b}})Sitaula, Aryal, Xiang, Basnet, and
  Lu}]{sitaula2020content}
Sitaula C, Aryal S, Xiang Y, Basnet A, Lu X (2021{\natexlab{b}}) Content and
  context features for scene image representation. Knowledge-Based Systems p
  107470

\bibitem[{Sitaula et~al(2020)Sitaula, Xiang, Basnet, Aryal, and
  Lu}]{sitaula2020hdf}
Sitaula C, Xiang Y, Basnet A, Aryal S, Lu X (2020) {HDF: Hybrid deep features
  for scene image representation}. In: Proc. Int. Joint Conf. on Neural
  Networks (IJCNN)

\bibitem[{Mikolov et~al(2013)Mikolov, Chen, Corrado, and
  Dean}]{mikolov2013efficient}
Mikolov T, Chen K, Corrado G, Dean J (2013) {Efficient estimation of word
  representations in vector space}. arXiv preprint arXiv:13013781

\bibitem[{Pennington et~al(2014)Pennington, Socher, and
  Manning}]{pennington2014glove}
Pennington J, Socher R, Manning C (2014) {Glove: Global vectors for word
  representation}. In: Proc. Conf. on Empirical Methods in Natural Language
  Processing (EMNLP), pp 1532--1543

\bibitem[{Bojanowski et~al(2017)Bojanowski, Grave, Joulin, and
  Mikolov}]{bojanowski2017enriching}
Bojanowski P, Grave E, Joulin A, Mikolov T (2017) {Enriching word vectors with
  subword information}. Trans of the Association for Computational Linguistics
  5:135--146

\bibitem[{Shahi and Sitaula(2021)}]{shahi2021natural}
Shahi TB, Sitaula C (2021) Natural language processing for nepali text: a
  review. Artificial Intelligence Review pp 1--29

\bibitem[{Nascimento et~al(2017)Nascimento, Laranjeira, Braz, Lacerda, and
  Nascimento}]{nascimento_robust_2017}
Nascimento G, Laranjeira C, Braz V, Lacerda A, Nascimento ER (2017) A robust
  indoor scene recognition method based on sparse representation. CoRR
  abs/1708.07555

\bibitem[{S{\'{a}}nchez et~al(2013)S{\'{a}}nchez, Perronnin, Mensink, and
  Verbeek}]{sanchez2013image}
S{\'{a}}nchez J, Perronnin F, Mensink T, Verbeek J (2013) {Image classification
  with the fisher vector: theory and practice}. Int J Comput Vis
  105(3):222--245

\bibitem[{Li et~al(2012)Li, Zhang, Guo, Bhanu, and An}]{li2012reference}
Li Q, Zhang H, Guo J, Bhanu B, An L (2012) Reference-based scheme combined with
  k-svd for scene image categorization. IEEE Signal Processing Letters
  20(1):67--70

\bibitem[{Ringn{\'e}r(2008)}]{ringner2008principal}
Ringn{\'e}r M (2008) What is principal component analysis? Nature biotechnology
  26(3):303--304

\bibitem[{Oliva and Torralba(2001)}]{oliva_modeling_2001}
Oliva A, Torralba A (2001) Modeling the shape of the scene: a holistic
  representation of the spatial envelope. Int J Comput Vis 42(3):145--175

\bibitem[{Zeglazi et~al(2016)Zeglazi, Amine, and Rziza}]{zeglazi_sift_2016}
Zeglazi O, Amine A, Rziza M (2016) Sift descriptors modeling and application in
  texture image classification. In: Proc. 13th Int. Conf. Comput. Graphics,
  Imaging and Visualization (CGiV), pp 265--268

\bibitem[{Wu and Rehg(2011)}]{wu_centrist:_2011}
Wu J, Rehg JM (2011) Centrist: a visual descriptor for scene categorization.
  IEEE Trans Pattern Anal Mach Intell 33(8):1489--1501

\bibitem[{Sinha et~al(2014)Sinha, Banerji, and Liu}]{sinha2014new}
Sinha A, Banerji S, Liu C (2014) New color gphog descriptors for object and
  scene image classification. Machine vision and applications 25(2):361--375

\bibitem[{Oliva(2005)}]{oliva2005Gist}
Oliva A (2005) {Gist of the scene}. In: Neurobiology of Attention, Elsevier, pp
  251--256

\bibitem[{Li et~al(2010)Li, Su, Fei-Fei, and Xing}]{li2010object}
Li LJ, Su H, Fei-Fei L, Xing EP (2010) {Object bank: A high-level image
  representation for scene classification {\&} semantic feature
  sparsification}. In: Proc. Adv. Neural Inf. Process. Syst. (NIPS), pp
  1378--1386

\bibitem[{Zhang et~al(2014)Zhang, Zhen, and Shao}]{zhang2014learning}
Zhang L, Zhen X, Shao L (2014) Learning object-to-class kernels for scene
  classification. IEEE Transactions on image processing 23(8):3241--3253

\bibitem[{Parizi et~al(2012)Parizi, Oberlin, and
  Felzenszwalb}]{parizi2012reconfigurable}
Parizi N, Oberlin JG, Felzenszwalb PF (2012) {Reconfigurable models for scene
  recognition}. In: Proc. Comput. Vis. Pattern Recognit.(CVPR), pp 2775--2782

\bibitem[{Juneja et~al(2013)Juneja, Vedaldi, Jawahar, and
  Zisserman}]{juneja2013blocks}
Juneja M, Vedaldi A, Jawahar C, Zisserman A (2013) {Blocks that shout:
  Distinctive parts for scene classification}. In: Proc. IEEE Conf. Comput.
  Vis. Pattern Recognit. (CVPR), pp 923--930

\bibitem[{Lin et~al(2014)Lin, Lu, Liao, and Jia}]{lin_learning_2014}
Lin D, Lu C, Liao R, Jia J (2014) Learning important spatial pooling regions
  for scene classification. In: Proc. IEEE Conf. Comput. Vis. Pattern Recognit.
  (CVPR), pp 3726--3733

\bibitem[{Quattoni and Torralba(2009)}]{quattoni_recognizing_2009}
Quattoni A, Torralba A (2009) {Recognizing indoor scenes}. In: Proc. IEEE Conf.
  Comput. Vis. Pattern Recognit. (CVPR), pp 413--420

\bibitem[{Zhu et~al(2010)Zhu, Li, Fei-Fei, and Xing}]{zhu_large_2010}
Zhu J, Li Lj, Fei-Fei L, Xing EP (2010) Large margin learning of upstream scene
  understanding models. In: Proc. Adv. Neural Inf. Process. Syst. (NIPS), pp
  2586--2594

\bibitem[{ShenghuaGao and Liang-TienChia(2010)}]{shenghuagao2010local}
ShenghuaGao IH, Liang-TienChia P (2010) {Local features are not
  lonely--Laplacian sparse coding for image classification}. In: Proc. IEEE
  Conf. Comput. Vis. Pattern Recognit. (CVPR), pp 3555--3561

\bibitem[{Perronnin et~al(2010)Perronnin, Sanchez, and
  Mensink}]{perronnin2010improving}
Perronnin F, Sanchez J, Mensink T (2010) {Improving the fisher kernel for
  large-scale image classification}. In: Proc. Eur. Conf. Comput. Vis. (ECCV),
  pp 143--156

\bibitem[{LeCun et~al(2015)LeCun, Bengio, and Hinton}]{lecun2015deep}
LeCun Y, Bengio Y, Hinton G (2015) Deep learning. nature 521(7553):436--444

\bibitem[{Sitaula et~al(2021)Sitaula, Basnet, Mainali, and
  Shahi}]{sitaula2021deep}
Sitaula C, Basnet A, Mainali A, Shahi T (2021) Deep learning-based methods for
  sentiment analysis on nepali covid-19-related tweets. Computational
  Intelligence and Neuroscience 2021

\bibitem[{Sitaula and Shahi(2022)}]{sitaula2022monkeypox}
Sitaula C, Shahi TB (2022) Monkeypox virus detection using pre-trained deep
  learning-based approaches. Journal of Medical Systems 46(11):1--9

\bibitem[{Shahi et~al(2022)Shahi, Sitaula, Neupane, and Guo}]{shahi2022fruit}
Shahi TB, Sitaula C, Neupane A, Guo W (2022) Fruit classification using
  attention-based mobilenetv2 for industrial applications. Plos one
  17(2):e0264,586

\bibitem[{He et~al(2016)He, Zhang, Ren, and Sun}]{he2016deep}
He K, Zhang X, Ren S, Sun J (2016) {Deep residual learning for image
  recognition}. In: Proc. IEEE Conf. Comput. Vis. Pattern Recognit. (CVPR), pp
  770--778

\bibitem[{Simonyan and Zisserman(2014)}]{simonyan2014very}
Simonyan K, Zisserman A (2014) {Very deep convolutional networks for
  large-scale image recognition}. arXiv preprint arXiv:14091556
  \eprint{1409.1556}

\bibitem[{Zhou et~al(2017)Zhou, Lapedriza, Khosla, Oliva, and
  Torralba}]{zhou2017places}
Zhou B, Lapedriza A, Khosla A, Oliva A, Torralba A (2017) {Places: A 10 million
  image database for scene recognition}. IEEE Trans Pattern Anal Mach Intell
  40(6):1452--1464

\bibitem[{Bai et~al(2019)Bai, Tang, and An}]{bai2019coordinate}
Bai S, Tang H, An S (2019) Coordinate cnns and lstms to categorize scene images
  with multi-views and multi-levels of abstraction. Expert Systems with
  Applications 120:298--309

\bibitem[{Krizhevsky et~al(2012)Krizhevsky, Sutskever, and
  Hinton}]{krizhevsky2012imagenet}
Krizhevsky A, Sutskever I, Hinton GE (2012) {Imagenet classification with deep
  convolutional neural networks}. In: Proc. Adv. Neural Inf. Process. Syst.
  (NIPS), pp 1097--1105

\bibitem[{Szegedy et~al(2014)Szegedy, Liu, Jia, Sermanet, Reed, Anguelov,
  Erhan, Vanhoucke, and Rabinovich}]{szegedy2015going}
Szegedy C, Liu W, Jia Y, Sermanet P, Reed S, Anguelov D, Erhan D, Vanhoucke V,
  Rabinovich A (2014) {Going deeper with convolutions}. In: Proc. IEEE Conf.
  Comput. Vis. Pattern Recognit. (CVPR), pp 1--9, \eprint{1409.4842}

\bibitem[{Gong et~al(2014)Gong, Wang, Guo, and
  Lazebnik}]{gong_multi-scale_2014}
Gong Y, Wang L, Guo R, Lazebnik S (2014) Multi-scale orderless pooling of deep
  convolutional activation features. In: Proc. Eur. Conf. Comput. Vis. (ECCV),
  pp 392--407

\bibitem[{Deng et~al(2009)Deng, Dong, Socher, Li, Li, and
  Fei-Fei}]{deng_imagenet:_2009}
Deng J, Dong W, Socher R, Li LJ, Li K, Fei-Fei L (2009) Imagenet: a large-scale
  hierarchical image database. In: Proc. IEEE Conf. Comput. Vis. Pattern
  Recognit. (CVPR)

\bibitem[{Zhou et~al(2016)Zhou, Khosla, Lapedriza, Torralba, and
  Oliva}]{zhou2016places}
Zhou B, Khosla A, Lapedriza A, Torralba A, Oliva A (2016) {Places: An image
  database for deep scene understanding}. arXiv preprint arXiv:161002055

\bibitem[{Jia et~al(2014)Jia, Shelhamer, Donahue, Karayev, Long, Girshick,
  Guadarrama, and Darrell}]{jia2014caffe}
Jia Y, Shelhamer E, Donahue J, Karayev S, Long J, Girshick R, Guadarrama S,
  Darrell T (2014) {Caffe: Convolutional architecture for fast feature
  embedding}. In: Proc. 22nd ACM Int. Conf. on Multimedia (ACMM), pp 675--678

\bibitem[{Kuzborskij et~al(2016)Kuzborskij, Maria~Carlucci, and
  Caputo}]{kuzborskij2016naive}
Kuzborskij I, Maria~Carlucci F, Caputo B (2016) {When naive bayes nearest
  neighbors meet convolutional neural networks}. In: Proc. IEEE Conf. Comput.
  Vis. Pattern Recognit. (CVPR), pp 2100--2109

\bibitem[{Fornoni and Caputo(2014)}]{fornoni2014scene}
Fornoni M, Caputo B (2014) Scene recognition with naive bayes non-linear
  learning. In: 2014 22nd International Conference on Pattern Recognition,
  IEEE, pp 3404--3409

\bibitem[{Tang et~al(2017)Tang, Wang, and Kwong}]{tang_g-ms2f:_2017}
Tang P, Wang H, Kwong S (2017) G-ms2f: Googlenet based multi-stage feature
  fusion of deep cnn for scene recognition. Neurocomputing 225:188--197

\bibitem[{Zhang et~al(2017)Zhang, Zhu, Huang, and Tian}]{zhang2017image}
Zhang C, Zhu G, Huang Q, Tian Q (2017) {Image classification by search with
  explicitly and implicitly semantic representations}. Information Sciences
  376:125--135

\bibitem[{Guo and Lew(2016)}]{guo2016bag}
Guo Y, Lew MS (2016) {Bag of Surrogate Parts: one inherent feature of deep
  cnns.} In: Proc. of the British Machine Vision Conference (BMVC)

\bibitem[{Gupta et~al(2021)Gupta, Dileep, and
  Thenkanidiyoor}]{gupta2021recognition}
Gupta S, Dileep AD, Thenkanidiyoor V (2021) Recognition of varying size scene
  images using semantic analysis of deep activation maps. Machine Vision and
  Applications 32(2):1--19

\bibitem[{Zhang et~al(2013)Zhang, Liu, Tian, Liang, and
  Huang}]{zhang2013beyond}
Zhang C, Liu J, Tian Q, Liang C, Huang Q (2013) Beyond visual features: A weak
  semantic image representation using exemplar classifiers for classification.
  Neurocomputing 120:318--324

\bibitem[{Yang and Ramanan(2015)}]{yang2015multi}
Yang S, Ramanan D (2015) {Multi-scale recognition with DAG-CNNs}. In: Proc.
  IEEE Int. Conf. Comput. Vis. (ICCV), pp 1215--1223

\bibitem[{Dixit et~al(2015)Dixit, Chen, Gao, Rasiwasia, and
  Vasconcelos}]{dixit2015scene}
Dixit M, Chen S, Gao D, Rasiwasia N, Vasconcelos N (2015) {Scene classification
  with semantic fisher vectors}. In: Proc. IEEE Conf. Comput. Vis. Pattern
  Recognit. (CVPR), pp 2974--2983

\bibitem[{Wang et~al(2017)Wang, Wang, Wang, Zhang, and Qiao}]{wang_weakly_2017}
Wang Z, Wang L, Wang Y, Zhang B, Qiao Y (2017) Weakly supervised patchnets:
  describing and aggregating local patches for scene recognition. IEEE Trans
  Image Process 26(4):2028--2041

\bibitem[{Guo et~al(2017)Guo, Huang, Wang, and Qiao}]{guo_locally_2017}
Guo S, Huang W, Wang L, Qiao Y (2017) Locally supervised deep hybrid model for
  scene recognition. IEEE Trans Image Process 26(2):808--820

\bibitem[{Khan et~al(2016)Khan, Hayat, Bennamoun, Togneri, and
  Sohel}]{khan_discriminative_2016}
Khan SH, Hayat M, Bennamoun M, Togneri R, Sohel FA (2016) A discriminative
  representation of convolutional features for indoor scene recognition. IEEE
  Trans Image Process 25(7):3372--3383

\bibitem[{Cheng et~al(2018)Cheng, Lu, Feng, Yuan, and Zhou}]{cheng_scene_2018}
Cheng X, Lu J, Feng J, Yuan B, Zhou J (2018) {Scene recognition with
  objectness}. Pattern Recognit 74:474--487

\bibitem[{Lin et~al(????)Lin, RoyChowdhury, and Maji}]{7968351}
Lin TYY, RoyChowdhury A, Maji S (????) Bilinear convolutional neural networks
  for fine-grained visual recognition. IEEE Trans Pattern Anal Mach Intell
  (6):1309--1322

\bibitem[{Jiang et~al(2019)Jiang, Chen, Song, and Liu}]{jiang2019deep}
Jiang S, Chen G, Song X, Liu L (2019) Deep patch representations with shared
  codebook for scene classification. ACM Trans on Multimedia Computing,
  Communications, and Applications 15(1s):1--17

\bibitem[{Sorkhi et~al(2020)Sorkhi, Hassanpour, and
  Fateh}]{sorkhi2020comprehensive}
Sorkhi AG, Hassanpour H, Fateh M (2020) A comprehensive system for image scene
  classification. Multimedia Tools and Applications pp 1--26

\bibitem[{Chen et~al(2020)Chen, Song, Zeng, and Jiang}]{chen2020scene}
Chen G, Song X, Zeng H, Jiang S (2020) Scene recognition with
  prototype-agnostic scene layout. IEEE Trans Image Processing 29:5877--5888

\bibitem[{Lopez-Cifuentes et~al(2020)Lopez-Cifuentes, Escudero-Vinolo, Bescos,
  and Garcia-Martin}]{lopez2020semantic}
Lopez-Cifuentes A, Escudero-Vinolo M, Bescos J, Garcia-Martin A (2020)
  Semantic-aware scene recognition. Pattern Recognit 102:107,256

\bibitem[{Zhang et~al(2020)Zhang, Wang, Lu, Wang, and Li}]{zhang2020locality}
Zhang B, Wang Q, Lu X, Wang F, Li P (2020) Locality-constrained affine subspace
  coding for image classification and retrieval. Pattern Recognit 100:107,167

\bibitem[{Wang and Mao(2019)}]{wang2019task}
Wang D, Mao K (2019) Task-generic semantic convolutional neural network for web
  text-aided image classification. Neurocomputing 329:103--115

\bibitem[{Kim(2014)}]{kim2014convolutional}
Kim Y (2014) {Convolutional neural networks for sentence classification}. arXiv
  preprint arXiv:14085882

\bibitem[{Bosch et~al(2008)Bosch, Zisserman, and Mu{\~n}oz}]{bosch2008scene}
Bosch A, Zisserman A, Mu{\~n}oz X (2008) Scene classification using a hybrid
  generative/discriminative approach. IEEE Trans Pattern Anal Mach Intell
  30(4):712--727

\bibitem[{Rasiwasia and Vasconcelos(2008)}]{rasiwasia2008scene}
Rasiwasia N, Vasconcelos N (2008) Scene classification with low-dimensional
  semantic spaces and weak supervision. In: IEEE Conf. Comput. Vis. Pattern
  Recognit. (CVPR), pp 1--6

\bibitem[{Van~Gemert et~al(2009)Van~Gemert, Veenman, Smeulders, and
  Geusebroek}]{van2009visual}
Van~Gemert JC, Veenman CJ, Smeulders AW, Geusebroek JM (2009) Visual word
  ambiguity. IEEE Trans Pattern Anal Mach Intell 32(7):1271--1283

\bibitem[{Zhang et~al(2014)Zhang, Cheng, Liu, Pang, Liang, Huang, and
  Tian}]{zhang2014object}
Zhang C, Cheng J, Liu J, Pang J, Liang C, Huang Q, Tian Q (2014) Object
  categorization in sub-semantic space. Neurocomputing 142:248--255

\bibitem[{Ali et~al(2018)Ali, Zafar, Riaz, Dar, Ratyal, Bajwa, Iqbal, and
  Sajid}]{ali2018hybrid}
Ali N, Zafar B, Riaz F, Dar SH, Ratyal NI, Bajwa KB, Iqbal MK, Sajid M (2018)
  {A hybrid geometric spatial image representation for scene classification}.
  PloS one 13(9):e0203,339

\bibitem[{Wang and Mao(2019)}]{wang2019learning}
Wang D, Mao K (2019) Learning semantic text features for web text-aided image
  classification. IEEE Trans Multimedia 21(12):2985--2996

\bibitem[{Griffin et~al(2007)Griffin, Holub, and Perona}]{griffin2007caltech}
Griffin G, Holub A, Perona P (2007) Caltech-256 object category dataset

\bibitem[{Wang et~al(2016)Wang, Wang, Wang, and Gao}]{wang2016csps}
Wang J, Wang W, Wang R, Gao W (2016) Csps: An adaptive pooling method for image
  classification. IEEE Transactions on Multimedia 18(6):1000--1010

\bibitem[{Sinha et~al(2012)Sinha, Banerji, and Liu}]{sinha2012novel}
Sinha A, Banerji S, Liu C (2012) Novel gabor-phog features for object and scene
  image classification. In: Joint IAPR International Workshops on Statistical
  Techniques in Pattern Recognition (SPR) and Structural and Syntactic Pattern
  Recognition (SSPR), Springer, pp 584--592

\bibitem[{Silberman and Fergus(2011)}]{silberman2011indoor}
Silberman N, Fergus R (2011) Indoor scene segmentation using a structured light
  sensor. In: 2011 IEEE international conference on computer vision workshops
  (ICCV workshops), IEEE, pp 601--608

\bibitem[{Ren et~al(2012)Ren, Bo, and Fox}]{ren2012rgb}
Ren X, Bo L, Fox D (2012) Rgb-(d) scene labeling: Features and algorithms. In:
  2012 IEEE Conference on Computer Vision and Pattern Recognition, IEEE, pp
  2759--2766

\bibitem[{Sun et~al(2018)Sun, Li, Liu, Han, and Wu}]{sun2018fusing}
Sun N, Li W, Liu J, Han G, Wu C (2018) Fusing object semantics and deep
  appearance features for scene recognition. IEEE Transactions on Circuits and
  Systems for Video Technology 29(6):1715--1728

\bibitem[{Redmon and Farhadi(2017)}]{redmon2017yolo9000}
Redmon J, Farhadi A (2017) Yolo9000: better, faster, stronger. In: Proceedings
  of the IEEE conference on computer vision and pattern recognition, pp
  7263--7271

\bibitem[{Liu and Tian(2019)}]{liu2019indoor}
Liu S, Tian G (2019) An indoor scene classification method for service robot
  based on cnn feature. Journal of Robotics 2019

\bibitem[{Liu et~al(2021)Liu, Tian, Zhang, and Duan}]{liu2021scene}
Liu S, Tian G, Zhang Y, Duan P (2021) Scene recognition mechanism for service
  robot adapting various families: A cnn-based approach using multi-type
  cameras. IEEE Transactions on Multimedia 24:2392--2406

\bibitem[{Choe et~al(2021)Choe, Seong, and Kim}]{choe2021indoor}
Choe S, Seong H, Kim E (2021) Indoor place category recognition for a cleaning
  robot by fusing a probabilistic approach and deep learning. IEEE Transactions
  on Cybernetics

\bibitem[{Fei-Fei and Perona(2005)}]{fei-fei_bayesian_2005}
Fei-Fei L, Perona P (2005) {A Bayesian hierarchical model for learning natural
  scene categories}. In: Proc. IEEE Comput. Soc. Conf. Comput. Vis. and Pattern
  Recognit. (CVPR), vol~2, pp 524--531

\bibitem[{Li and Li(2007)}]{li2007and}
Li LJ, Li FF (2007) {What, where and who? classifying events by scene and
  object recognition.} In: Proc. 11th Int. Conf. Comput. Vis. (ICCV), vol~2,
  p~6

\bibitem[{Xiao et~al(2010)Xiao, Hays, Ehinger, Oliva, and
  Torralba}]{xiao2010sun}
Xiao J, Hays J, Ehinger KA, Oliva A, Torralba A (2010) {Sun database:
  Large-scale scene recognition from abbey to zoo}. In: Proc. IEEE Conf.
  Comput. Vis. Pattern Recognit. (CVPR), pp 3485--3492

\bibitem[{Niu et~al(2010)Niu, Zhou, and Shi}]{niu2010hybrid}
Niu Z, Zhou Y, Shi K (2010) A hybrid image representation for indoor scene
  classification. In: 2010 25th International Conference of Image and Vision
  Computing New Zealand, IEEE, pp 1--7

\bibitem[{Cho and Lam(2012)}]{cho2012efficient}
Cho WS, Lam KM (2012) An efficient and effective hybrid pyramid kernel for
  un-segmented image classification. In: 2012 International Conference on
  Systems and Informatics (ICSAI2012), IEEE, pp 2153--2158

\bibitem[{Chen et~al(2018)Chen, Xie, Wang, and Zhao}]{chen2018scene-hillc}
Chen H, Xie K, Wang H, Zhao C (2018) Scene image classification using
  locality-constrained linear coding based on histogram intersection.
  Multimedia Tools and Applications 77(3):4081--4092

\bibitem[{Banerji et~al(2012)Banerji, Sinha, and Liu}]{banerji2012novel}
Banerji S, Sinha A, Liu C (2012) Novel color, shape and texture-based scene
  image descriptors. In: 2012 IEEE 8th International Conference on Intelligent
  Computer Communication and Processing, IEEE, pp 245--248

\bibitem[{Li et~al(2013)Li, Qin, Chai, Zhang, Guo, and
  Bhanu}]{li2013representative}
Li Q, Qin Z, Chai L, Zhang H, Guo J, Bhanu B (2013) Representative
  reference-set and betweenness centrality for scene image categorization. In:
  2013 IEEE International Conference on Image Processing, IEEE, pp 3254--3258

\bibitem[{Sinha et~al(2014)Sinha, Banerji, and Liu}]{sinha2014scene}
Sinha A, Banerji S, Liu C (2014) Scene image classification using a
  wigner-based local binary patterns descriptor. In: 2014 International Joint
  Conference on Neural Networks (IJCNN), IEEE, pp 1614--1621

\bibitem[{Hu and Guo(2012)}]{hu2012spatial}
Hu J, Guo P (2012) Spatial local binary patterns for scene image
  classification. In: 2012 6th International Conference on Sciences of
  Electronics, Technologies of Information and Telecommunications (SETIT),
  IEEE, pp 326--330

\bibitem[{Zuo et~al(2015)Zuo, Wang, Shuai, Zhao, and Yang}]{zuo2015exemplar}
Zuo Z, Wang G, Shuai B, Zhao L, Yang Q (2015) Exemplar based deep
  discriminative and shareable feature learning for scene image classification.
  Pattern Recognition 48(10):3004--3015

\bibitem[{Liu et~al(2019)Liu, Tian, and Xu}]{liu2019novel}
Liu S, Tian G, Xu Y (2019) A novel scene classification model combining resnet
  based transfer learning and data augmentation with a filter. Neurocomputing
  338:191--206

\bibitem[{Khan et~al(2021)Khan, Chefranov, and Demirel}]{khan2021image}
Khan A, Chefranov A, Demirel H (2021) Image scene geometry recognition using
  low-level features fusion at multi-layer deep cnn. Neurocomputing
  440:111--126

\bibitem[{Liu et~al(2018)Liu, Li, and Wu}]{liu2018attribute}
Liu W, Li Y, Wu Q (2018) An attribute-based high-level image representation for
  scene classification. IEEE Access 7:4629--4640

\bibitem[{Qi and Wang(2016)}]{qi2016cssr}
Qi M, Wang Y (2016) Deep-cssr: Scene classification using category-specific
  salient region with deep features. In: 2016 IEEE International Conference on
  Image Processing (ICIP), IEEE, pp 1047--1051

\bibitem[{Xie et~al(2018)Xie, Jin, Zhang, Zang, Yang, Wang, and
  Pu}]{xie2018class}
Xie GS, Jin XB, Zhang XY, Zang SF, Yang C, Wang Z, Pu J (2018) From
  class-specific to class-mixture: Cascaded feature representations via
  restricted boltzmann machine learning. IEEE Access 6:69,393--69,406

\bibitem[{Bai(2017)}]{bai2017growing}
Bai S (2017) Growing random forest on deep convolutional neural networks for
  scene categorization. Expert systems with applications 71:279--287

\bibitem[{Sharma et~al(2018)Sharma, Gupta, Dileep, and
  Rameshan}]{sharma2018scene}
Sharma K, Gupta S, Dileep AD, Rameshan R (2018) Scene image classification
  using reduced virtual feature representation in sparse framework. In: 2018
  IEEE International Conference on Acoustics, Speech and Signal Processing
  (ICASSP), IEEE, pp 2701--2705

\bibitem[{Dixit et~al(2019)Dixit, Li, and Vasconcelos}]{dixit2019semantic}
Dixit M, Li Y, Vasconcelos N (2019) Semantic fisher scores for task transfer:
  Using objects to classify scenes. IEEE transactions on pattern analysis and
  machine intelligence 42(12):3102--3118

\bibitem[{Lin et~al(2021)Lin, Lee, Cai, Chen, and Chen}]{lin2021global}
Lin C, Lee F, Cai J, Chen H, Chen Q (2021) Global and graph encoded local
  discriminative region representation for scene recognition. Computer Modeling
  in Engineering \& Sciences 128(3):985--1006

\bibitem[{Wu et~al(2015)Wu, Wang, Wang, and Yu}]{wu2015harvesting}
Wu R, Wang B, Wang W, Yu Y (2015) Harvesting discriminative meta objects with
  deep cnn features for scene classification. In: Proceedings of the IEEE
  International Conference on Computer Vision, pp 1287--1295

\bibitem[{Wang et~al(2022)Wang, Peng, and De~Baets}]{wang2022joint}
Wang C, Peng G, De~Baets B (2022) Joint global metric learning and local
  manifold preservation for scene recognition. Information Sciences
  610:938--956

\bibitem[{Aria and Cuccurullo(2017)}]{aria2017bibliometrix}
Aria M, Cuccurullo C (2017) bibliometrix: An r-tool for comprehensive science
  mapping analysis. Journal of informetrics 11(4):959--975

\end{thebibliography}
\end{document}